\documentclass[11pt]{article}

\usepackage[margin=1in]{geometry}
\usepackage{amsmath,amssymb,amsthm}
\usepackage{graphicx}
\usepackage{bm}
\usepackage{hyperref}
\hypersetup{hidelinks}
\usepackage{cleveref}
\usepackage{xcolor}
\usepackage{booktabs}

\newcommand{\R}{\mathbb{R}}
\newcommand{\E}{\mathbb{E}}
\newcommand{\tr}{\operatorname{tr}}
\newcommand{\diag}{\operatorname{diag}}

\title{The Dynamics of Spectral Bias in Operator Learning}

\author{%
Conrad Ainslie\\
\small Department of Applied Mathematics,\\[-2pt]
\small University of California, Santa Cruz, USA
\and
Ching-Yao Lai\thanks{Corresponding author: \texttt{cyaolai@stanford.edu}}\\
\small Department of Geophysics and the Institute for Computational\\[-2pt]
\small and Mathematical Engineering, Stanford University, USA
\and
Ashesh Chattopadhyay\thanks{Corresponding author: \texttt{aschatto@ucsc.edu}}\\
\small Department of Applied Mathematics,\\[-2pt]
\small University of California, Santa Cruz, USA
}

\date{}

\begin{document}
\maketitle

\begin{abstract}
Spectral bias is the phenomenon where neural networks learn the low-wavenumber content of a target before its high-wavenumber content. We analyze the training dynamics of neural operators on the discrete Poisson problem and find that the bias is a property of the eigenbasis of the kernel that governs the training update, not of the wavenumber basis in which it is usually reported. A model learning a discretized operator retains its error in the eigendirections of this kernel with small eigenvalues, which are the rates at which training contracts the error along those directions. Under smooth forcing, low variance is placed at high wavenumbers, and any model, linear or nonlinear, has small eigenvalues along the eigenvectors with high-wavenumber content. Under white forcing, the linear models learn the operator, while the nonlinear models retain error in slow directions that depend on the coupling between training samples in the kernel; for the GELU dense network they disappear entirely when that coupling is removed. A model whose representation omits a set of modes has a zero eigenvalue for each of them. What the retained error looks like in wavenumber then depends on how these eigendirections project onto the sine modes. When those eigendirections are the sine modes, as for a spectral operator or a linear Fourier neural operator, each wavenumber has its own rate and the error concentrates at the high wavenumbers. When they are delocalized, as for dense networks under both forcings, the error has the same magnitude at every wavenumber. The spectrum of the true solution falls with wavenumber, so the relative error grows with wavenumber once the true solution drops below the error, and spectral bias appears at the high wavenumbers without any mode being learned more slowly than another. The eigenvalues are responsible for how much error remains and the eigenvectors govern where it appears in wavenumber. The high-wavenumber bias usually reported is this retained error seen in the sine basis.
\end{abstract}

% ===============================================================
\section{Introduction}
\label{sec:intro}
% ===============================================================

Neural networks trained on a target with structure at many scales do not fit all scales at the same rate. In most reported cases, the low-wavenumber content of the target is learned before the high-wavenumber content~\cite{rahaman2019spectral,xu2020frequency,basri2019convergence,tancik2020fourier}, and this behavior is called spectral bias. In autoregressive emulators of turbulent flows, the small scales that spectral bias leaves unlearned are amplified during time integration and make long-term emulation unstable~\cite{chattopadhyay2023challenges}. Classical iterative solvers for elliptic equations show the opposite ordering, with oscillatory error damped within a few iterations while smooth error persists~\cite{briggs2000multigrid}.

The explanation usually given for spectral bias is a per-mode learning rate, under which each wavenumber is learned at its own rate and the high wavenumbers are learned slowly. Kernel analyses state the same thing in the eigenbasis of the neural tangent kernel~\cite{jacot2018ntk,bordelon2020spectrum,canatar2021spectral}. For uniformly sampled inputs in low dimension the eigenfunctions of that kernel are Fourier modes~\cite{basri2019convergence,bietti2019inductive,cao2021towards}, so the two descriptions coincide. In this paper we show that, in operator learning~\cite{lu2021learning,li2021fourier,kovachki2023neural}, the per-mode description holds only when the eigenvectors of the operator that governs the training update are aligned with the sine basis. For generic dense networks they are not. The trained error of these networks is nearly flat across wavenumber, at a level determined by the eigenvalues of that operator, and the relative error grows with wavenumber because the spectrum of the true solution decays through this flat floor, not because the network learns high wavenumbers more slowly.

Several kernel descriptions of spectral bias in scientific machine learning precede this work. Wang et al.~\cite{wang2022when,wang2021understanding} derive the neural tangent kernel (NTK) of physics-informed neural networks~\cite{raissi2019physics}, show that it is constant during training in the infinite-width limit under gradient descent, and trace training failure to different convergence rates of the loss components, determined by the eigenvalues of that kernel. Wang, Wang and Perdikaris~\cite{wang2021eigenvector} identify spectral bias with a bias toward the leading eigenvectors of the limiting NTK, and show that Fourier feature embeddings shift these eigenvectors to higher frequencies. Seroussi et al.~\cite{seroussi2024spectral} analyze over-parameterized networks through an equivalence with Gaussian process regression and describe spectral bias as a statement about kernel--task alignment. Khodakarami et al.~\cite{khodakarami2026spectral} report wavenumber-resolved error metrics across physics-informed and operator-learning settings and find that spectral bias depends on the optimizer as well as on the architecture. For neural operators, Khodakarami et al.~\cite{khodakarami2026mitigating} reduce it by scaling the high-frequency components of the latent features. Xu et al.~\cite{xu2025spectral} review mitigation strategies, mainly frequency shifting and scaling, that assume a low-wavenumber-first learning order. Wang and Lai~\cite{wang2024multistage} remove the error plateau of regression and physics-informed networks, near $10^{-5}$, by training successive networks on the residue of the previous one, whose content moves to higher frequencies at each stage.

Our analysis differs from these in three respects. First, the operator analyzed is the kernel of the update actually applied. The models are trained with Adam, so the kernel carries the diagonal preconditioner of the optimizer (Section~\ref{sec:kernel-bias}), and the gradient-flow NTK of~\cite{jacot2018ntk,wang2022when} is recovered when the preconditioner is the identity. Second, the batch average of minibatch training is kept inside the kernel, so the kernel contains cross terms that couple every sample's error to every other sample's error through the shared parameters. Section~\ref{sec:error-loading} shows that for the dense networks the eigendirections retaining the trained error exist only through these cross terms, so an analysis that drops them cannot locate the error. Third, we make no infinite-width or constant-kernel assumption; outside the lazy regime the kernel evolves during training~\cite{chizat2019lazy,fort2020deep,bordelon2022self}. All kernel statistics are computed over the final steps of training, so they relate the trained error to the operator that drives the update at that stage, not to a rate law fixed at initialization.

We develop this result for the discrete Poisson problem, using a hierarchy of five operator models that ranges from an ideal spectral operator to unstructured dense networks (Section~\ref{sec:setup}). Section~\ref{sec:linear-op} treats the linear case, where the operator that governs the update is the covariance of the training data and the learning order is determined exactly by its spectrum. Section~\ref{sec:nonlinear} treats the nonlinear case, where the operator that governs the update is the stacked kernel of the Adam iteration. We show that every model that fails to learn the operator retains its error in the slow eigendirections of this kernel, that the slow rates arise from the data, from the nonlinearity through the coupling between training samples, or from the representation, and that the shape of the retained error in the sine basis is determined by how these eigendirections project onto the sine modes, from a per-mode learning rate when they are sine-aligned to a flat floor when they are delocalized. Section~\ref{sec:discussion} discusses the results. The scalar regression case, in which the same statements can be made in closed form, is treated in Appendix~\ref{sec:scalar}.

% ===============================================================
\section{Operator learning for the Poisson problem}
\label{sec:setup}
% ===============================================================
The operator learned here is the inverse of the one-dimensional discrete Poisson operator. The input is the forcing, discretized as a vector $f \in \R^N$, and the target is the solution vector
\begin{equation}
u = A^{-1} f \in \R^N,
\label{eq:target-op}
\end{equation}
where $A \in \R^{N \times N}$ is the discrete Poisson operator. In scalar regression, spectral bias is usually analyzed through the Fourier coefficients of a single target function. Here the unknown object is a linear map.

% ---------------------------------------------------------------
\subsection{Discrete Poisson problem and the target operator}
\label{sec:poisson-setup}
% ---------------------------------------------------------------

Consider the Poisson equation on $(0,1)$ with homogeneous Dirichlet boundary conditions,
\begin{equation}
-u_{xx}(x) = f(x), \qquad u(0) = u(1) = 0.
\label{eq:poisson}
\end{equation}
A second-order finite-difference discretization on $N$ interior grid points gives $A u = f$ with
\begin{equation}
A = \frac{1}{h^2}
\begin{pmatrix}
2 & -1 & & \\
-1 & 2 & \ddots & \\
& \ddots & \ddots & -1 \\
& & -1 & 2
\end{pmatrix},
\text{where} \quad h = \frac{1}{N+1}.
\label{eq:A-matrix}
\end{equation}
The natural basis for operator $A$ is the orthonormal discrete sine basis. Let $S \in \R^{N \times N}$ have entries
\begin{equation}
S_{jk} = \sqrt{\frac{2}{N+1}} \sin\!\left( \frac{\pi j k}{N+1}
\right),
\qquad j, k = 1, \dots, N,
\label{eq:sine-basis}
\end{equation}
with columns $v_k$, $k = 1, \dots, N$. Then $A = S \Lambda_A S^\top$ with
\begin{equation}
\Lambda_A = \diag(\kappa_1, \dots, \kappa_N),
\qquad
\kappa_k = \frac{4}{h^2} \sin^2\!\left( \frac{\pi k}{2(N+1)}
\right).
\label{eq:poisson-eigs}
\end{equation}
In the modal coordinates $\hat{f} = S^\top f$ and $\hat{u} = S^\top u$ the solution is
\begin{equation}
\hat{u}_k = \frac{1}{\kappa_k}\, \hat{f}_k,
\label{eq:modal-gain}
\end{equation}
so the target operator is diagonal in the sine basis with gain $g^{*}_k = 1/\kappa_k \sim k^{-2}$ for small $k$. The inverse Poisson operator is low-pass, and high-wavenumber forcing is attenuated in the solution. This is a property of the map being learned. It must be kept separate from any wavenumber dependence of the learning process, which is the subject of this paper.

\paragraph{Forcing ensembles.}
Training data are generated by sampling forcings with independent zero-mean modal coefficients, $f = \sum_k \hat{f}_k v_k$ with $\E[\hat{f}_k \hat{f}_j] = \sigma_k^2 \delta_{kj}$, so the forcing covariance $\Sigma_f := \E[f f^\top]$ is diagonal in the sine basis with eigenvalues $\sigma_k^2$. We use two ensembles. \textit{Smooth forcing} has $\sigma_k^2 = k^{-4}$ and concentrates variance at low $k$. \textit{White forcing} has $\sigma_k^2 = 1$ and excites all modes equally.

\paragraph{Architecture hierarchy.}
We train five operator models. The ideal diagonal spectral operator has one trainable gain for each of its $K_{\mathrm{ret}} = 32$ retained sine modes and no other parameters. The linear FNO and the nonlinear FNO~\cite{li2021fourier,kovachki2023neural} are Fourier neural operators whose layers combine per-mode spectral weights with pointwise channel mixing, without and with a GELU nonlinearity~\cite{hendrycks2016gelu} respectively. The GELU dense network is a multilayer perceptron acting directly on the grid values. The deep linear dense network is the same multilayer perceptron without nonlinearities or biases, so its product of weight matrices realizes a single linear operator. We refer to the last four as the generic models and to the last two as the dense networks. All models are trained with Adam~\cite{kingma2015adam} on the mean-squared error over freshly sampled forcing batches. Architecture and training details are given in Appendix~\ref{app:details}.

% ---------------------------------------------------------------
\subsection{A flat error floor}
\label{sec:phenomenon}
% ---------------------------------------------------------------

\begin{figure}[tp]
\centering
\includegraphics[width=\textwidth]{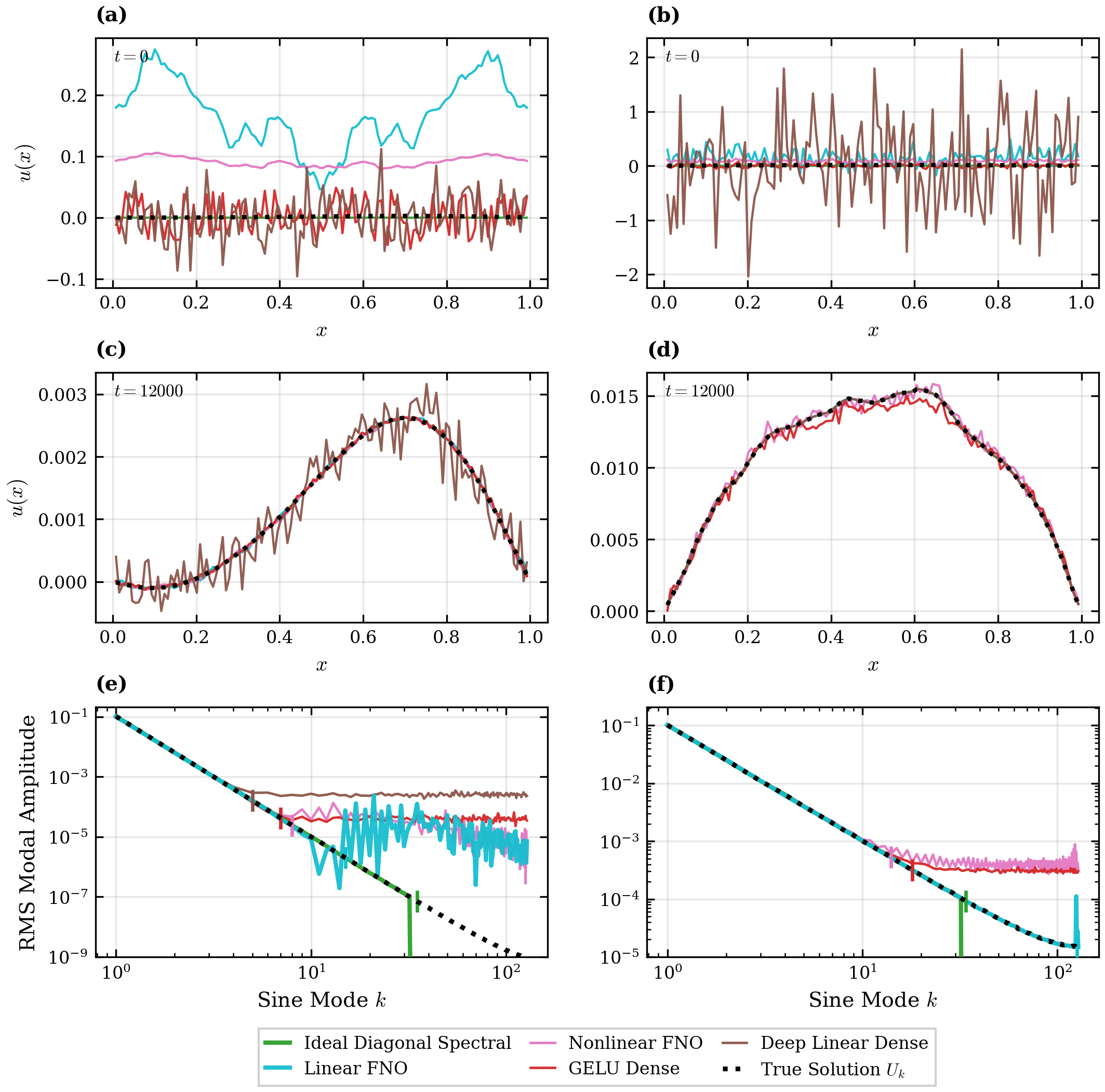}
\caption{\textbf{Spectral bias in the trained hierarchy} (smooth forcing left, white forcing right). (a,b)~Model outputs for one validation forcing at the first training step ($t = 0$) against the true solution (black). (c,d)~The same at the final training step ($t = 12000$). (e,f)~RMS modal amplitude of the trained outputs against the true solution spectrum $U_k$ of Eq.~(\ref{eq:true-spectrum}). Vertical ticks mark the mode at which each model departs from $U_k$, and the crossing, computed after a three-point median filter on $E_k$, defines the gap mode $k_{\mathrm{gap}}$ of Eq.~(\ref{eq:kgap}). The spike of the linear FNO at the end of the band in (f) is an input-independent offset at the highest odd modes (Appendix~\ref{app:parity}).}
\label{fig:spectral-gap}
\end{figure}

For a model $u_\theta = F_\theta(f)$ we define the RMS modal amplitude of the output over a validation set of forcings indexed by $\mu$, $\langle (S^\top u_\theta^\mu)_k^2 \rangle_\mu^{1/2}$, and the spectrum of the true solutions,
\begin{equation}
U_k = \big\langle \big(S^\top u^\mu\big)_k^2 \big\rangle_\mu^{1/2}.
\label{eq:true-spectrum}
\end{equation}
Figure~\ref{fig:spectral-gap} shows the trained hierarchy in physical space and in the sine basis. In physical space, the trained outputs overlay the true solution up to a fine-scale residual in Fig.~\ref{fig:spectral-gap}(c,d). In the sine basis, each output tracks $U_k$ up to a model-dependent mode. Beyond it, the generic models stay near an error floor, flat for the dense networks and almost flat for the nonlinear FNO, while $U_k$ continues to decay, and the ideal diagonal spectral operator drops to zero at its truncation in Fig.~\ref{fig:spectral-gap}(e,f). Under white forcing the linear FNO and the deep linear dense network track $U_k$ over the whole band. The mode at which each model departs from the target depends on the architecture and the forcing, and training lowers the level of the flat portion without changing its shape. Relative to the target, the error therefore grows with $k$ beyond the departure mode. \textit{This is the phenomenon usually reported as spectral bias}.

The same observation can be stated as an error spectrum. For a trained model we define the RMS modal error over the validation set,
\begin{equation}
E_k
= \big\langle \big(S^\top (u_\theta^\mu - u^\mu)\big)_k^2
\big\rangle_\mu^{1/2},
\label{eq:abs-modal-error}
\end{equation}
and the relative RMS modal error,
\begin{equation}
\varepsilon_k
= \sqrt{\frac{\big\langle \big(S^\top (u_\theta^\mu -
u^\mu)\big)_k^2 \big\rangle_\mu}
{\big\langle \big(S^\top u^\mu\big)_k^2 \big\rangle_\mu}},
\label{eq:rel-modal-error}
\end{equation}
so that $\varepsilon_k = E_k / U_k$.

\begin{figure}[tp]
\centering
\includegraphics[width=\textwidth]{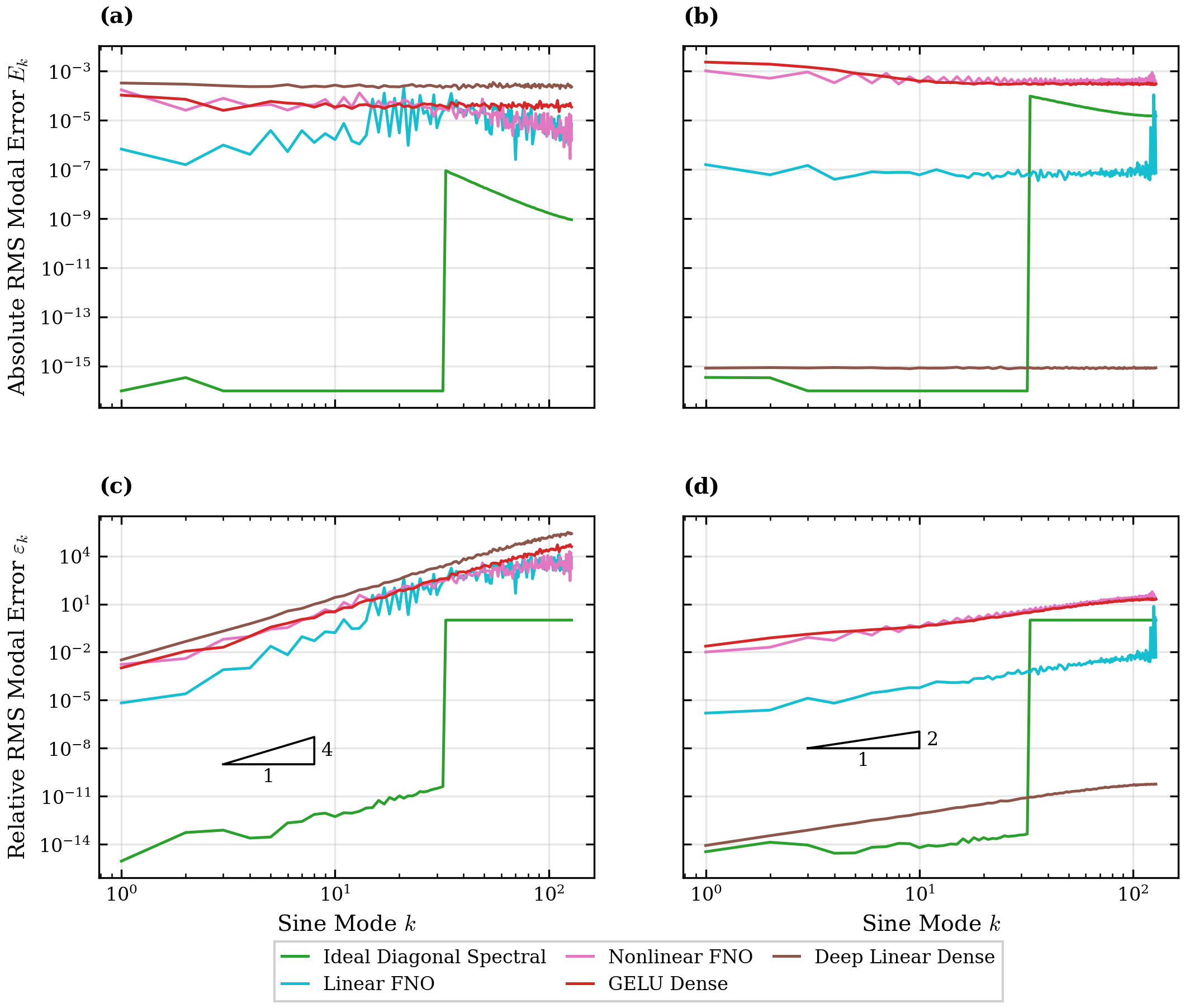}
\caption{\textbf{Final modal error spectra of the hierarchy} (smooth forcing left, white forcing right). (a,b)~Modal error $E_k$ of Eq.~(\ref{eq:abs-modal-error}). (c,d)~Relative modal error $\varepsilon_k$ of Eq.~(\ref{eq:rel-modal-error}). Slope triangles mark $\varepsilon_k \propto k^4$ (smooth) and $k^2$ (white), the growth predicted for a flat $E_k$ and $k \ll N$.}
\label{fig:final-spectra}
\end{figure}

Figure~\ref{fig:final-spectra} shows both error spectra at the end of training for all five models. Beyond a low-$k$ regime, the modal error $E_k$ of the dense networks and of the nonlinear FNO is nearly flat in $k$ for both forcings in Fig.~\ref{fig:final-spectra}(a,b), at a level determined by the model and the forcing. The linear FNO is flat at a low level under white forcing but rises with $k$ under smooth forcing, and the ideal diagonal spectral operator learns its retained modes exactly and increases after the truncation. Section~\ref{sec:fourier-gap} explains this range of shapes by the projection of the slow eigendirections of the training kernel onto the sine basis. The relative modal error $\varepsilon_k$ in Fig.~\ref{fig:final-spectra}(c,d) grows with $k$ for every generic model under smooth forcing, which is the well-known signature of spectral bias. The FNO models, however, level off at high $k$. Under smooth forcing their eigenvectors are localized on one or a few sine modes in Fig.~\ref{fig:misalignment}(a,c), more cleanly than under white forcing in Fig.~\ref{fig:misalignment-white}(a,c), so each mode approximately keeps its own error, $\sigma_k$ times its gain error, which falls with $k$ in Fig.~\ref{fig:final-spectra}(a). The apparent severity of spectral bias decreases as the forcing places more energy in the high modes: $\varepsilon_k$ grows much more slowly under white forcing in Fig.~\ref{fig:final-spectra}(d) than under smooth forcing in Fig.~\ref{fig:final-spectra}(c), consistent with the linear analysis of Section~\ref{sec:linear-op}.

The growth of $\varepsilon_k$ does not label a mode at which learning dynamics change. For the dense networks and the nonlinear FNO, $E_k$ is nearly constant in $k$ beyond a model-dependent mode in Fig.~\ref{fig:final-spectra}(a,b) (see also Section~\ref{sec:fourier-gap}), so the growth of $\varepsilon_k = E_k / U_k$ is the decay of its denominator. There is a single error level and a mode beyond which the target falls below it. We call this crossing the gap mode,
\begin{equation}
k_{\mathrm{gap}} := \min \left\{ k : E_k \geq U_k
\right\}.
\label{eq:kgap}
\end{equation}
For $k \geq k_{\mathrm{gap}}$ the output at mode $k$ is dominated by error and stays at the floor regardless of the true solution at that mode. When $E_k < U_k$ at every mode, as for the deep linear dense network under white forcing, there is no gap. This is why the output spectra of Fig.~\ref{fig:spectral-gap}(e,f) are flat in the gap region while the target continues to decay. Section~\ref{sec:nonlinear} traces the level of the floor to the operator that governs the training update.

% ===============================================================
\section{The data covariance governs the learning order in linear operator learning}
\label{sec:linear-op}
% ===============================================================

Under a linear learning rule, the error $e$ about the minimizer evolves as $\dot{e} = -\mathcal{T} e$, where $\mathcal{T}$ is the symmetric operator representing the update. Expanding $e$ in the eigenpairs $(\phi_\alpha, \lambda_\alpha)$ of $\mathcal{T}$ gives
\begin{equation}
e(t) = \sum_\alpha e^{-\lambda_\alpha t}\, a_\alpha(0)\,
\phi_\alpha,
\qquad
a_\alpha(0) = \phi_\alpha^\top e(0).
\label{eq:modal-law}
\end{equation}
A discrete iteration with step size $\eta$ contracts the same coefficients by the factor $1 - \eta \lambda_\alpha$ per step, which is the form taken by the scalar models of Appendix~\ref{sec:scalar}. The rates are determined entirely by the eigenvalues $\lambda_\alpha$. Which mode is learned first depends on which operator appears in the update, and any statement about wavenumber-dynamics of the error is a statement about how the eigenvectors of that operator project onto the sine basis.

Let the model be a single matrix, $u_\theta = M f$ with $M \in \R^{N \times N}$, trained on the population mean-squared error. With the operator error $\Delta := M - A^{-1}$, the loss is
\begin{equation}
\mathcal{L}(M) = \frac{1}{2}\, \E \| M f - A^{-1} f \|^2
= \frac{1}{2}\, \E\!\left[ f^\top \Delta^\top \Delta\, f \right]
= \frac{1}{2} \tr\!\left( \Delta\, \Sigma_f\, \Delta^\top \right),
\label{eq:op-loss}
\end{equation}
where the last equality uses the cyclic property of the trace. The training distribution of the forcing functions appears in the loss through its covariance $\Sigma_f$. The gradient is $\nabla_M \mathcal{L} = \Delta \Sigma_f$, and under gradient flow, the operator error obeys
\begin{equation}
\dot{\Delta} = -\Delta\, \Sigma_f,
\qquad
\Delta(t) = \Delta(0)\, e^{-\Sigma_f t}.
\label{eq:linear-op-law}
\end{equation}
The initial operator error is damped exponentially in the directions selected by $\Sigma_f$. This is Eq.~(\ref{eq:modal-law}) with the forcing covariance as the update operator, acting on each row of $\Delta$ from the right. Writing
\begin{equation}
\Sigma_f = Q\, \Lambda_f\, Q^\top,
\qquad
\Lambda_f = \diag(\sigma_1^2, \dots, \sigma_N^2),
\label{eq:sigma-eig}
\end{equation}
with orthonormal eigenvectors $q_k$, and projecting onto $q_k$,
\begin{equation}
\Delta(t)\, q_k = e^{-\sigma_k^2 t}\, \Delta(0)\, q_k.
\label{eq:linear-op-rates}
\end{equation}
The learning order in shallow linear operator learning is governed by the forcing spectrum $\{\sigma_k^2\}$, not by the spectrum of the solution operator in the Poisson equation~$\{\kappa_k\}$. A direction with large variance is excited often by the training data, so error along that direction is penalized strongly and decays fast. A direction with low variance is rarely excited and its error persists. For the forcings of Section~\ref{sec:poisson-setup}, the covariance is diagonal in the sine basis, $q_k = v_k$ with $v_k$ being the $k$th column of $S$, and Eq.~(\ref{eq:linear-op-rates}) becomes
\begin{equation}
\Delta(t)\, v_k = e^{-\sigma_k^2 t}\, \Delta(0)\, v_k.
\label{eq:linear-op-sine}
\end{equation}
Under white forcing all modes are learned at the same rate. Under smooth forcing the low modes are learned first and the high-mode error is nearly untouched.

\paragraph{Loss decomposition in the sine basis.}
The role of the covariance is clearest when the learned operator is itself diagonal in the sine basis, $(S^\top u_\theta)_k = g_k \hat{f}_k$. The per-sample modal error is then $(g_k - 1/\kappa_k)\hat{f}_k$ and the population loss decomposes mode by mode,
\begin{equation}
\mathcal{L} = \frac{1}{2} \sum_{k=1}^{N} \sigma_k^2
\left( g_k - \frac{1}{\kappa_k} \right)^{\!2}.
\label{eq:loss-decomp}
\end{equation}
The gain error at mode $k$ is weighted by the variance of that mode in the forcing. Under white forcing every mode contributes comparably to the loss. Under smooth forcing the high-mode gain errors are weakly penalized, and a model can reach a small total loss while remaining inaccurate at large $k$.

\paragraph{Physical damping versus learning order.}
The low-pass structure of the target operator must be kept separate from the learning order. The gain $g^{*}_k \sim k^{-2}$ means that, even under white forcing, the true solution has less high-wavenumber energy than the forcing. This is a property of the inverse Poisson operator. The learning dynamics of Eq.~(\ref{eq:linear-op-law}) are controlled by $\Sigma_f$, not by $A$. If low wavenumbers are learned first in this linear setting, it is because the training data place more variance there, and not because the Poisson operator smooths the optimization. This is also different from Jacobi iteration, where the damping rates are eigenvalues of an iteration matrix built from $A$ itself (Section~\ref{sec:discussion}). Figures~\ref{fig:spectral-gap} and~\ref{fig:final-spectra} show the effect of the data directly. Under white forcing the linear models have lower error and no accuracy gap, even though white forcing excites every mode. The deep linear dense network lies on the true solution in Fig.~\ref{fig:spectral-gap}(d,f), and its modal error is at machine epsilon in Fig.~\ref{fig:final-spectra}(b). This is the benchmark against which the nonlinear results of Section~\ref{sec:nonlinear} should be compared. In the linear case the update operator is the data covariance, and spectral bias is exactly the data spectrum. The forcing spectrum is the first of three sources of slow rates identified in this paper. Section~\ref{sec:nonlinear} adds the nonlinearity of the network and the truncation of the representation.

% ===============================================================
\section{Nonlinear operator learning}

\label{sec:nonlinear}

% ---------------------------------------------------------------
\subsection{The stacked kernel of the Adam update}
\label{sec:kernel-bias}
% ---------------------------------------------------------------

A neural operator is a map $F_\theta : \R^N \to \R^N$, $u_\theta = F_\theta(f)$, with parameters $\theta \in \R^P$. Given a training batch $\{(f^\mu, u^\mu)\}_{\mu=1}^{B}$ with $u^\mu = A^{-1} f^\mu$ and per-sample errors $e^\mu(\theta) = F_\theta(f^\mu) - u^\mu$ (sample indices are superscripts on vectors and subscripts on matrices throughout), the batch gradient of the mean-squared error, the mean over the $BN$ entries of the batch, is $\tfrac{2}{BN} \sum_\mu J_\mu^\top e^\mu$ with $J_\mu := \partial F_\theta(f^\mu) / \partial \theta$. The models are trained with Adam, which we model as gradient descent preconditioned by the diagonal matrix
\begin{equation}
\Pi = \diag\!\Big( \big(s_1^{1/2} +
\epsilon_{\mathrm{A}}\big)^{-1}, \dots,
\big(s_P^{1/2} + \epsilon_{\mathrm{A}}\big)^{-1} \Big),
\label{eq:adam-precond}
\end{equation}
where $s_i$ is the running second-moment estimate of gradient coordinate $i$ and $\epsilon_{\mathrm{A}}$ is the stabilization constant. We neglect the momentum average. Appendix~\ref{app:details} shows that this is valid for the slow eigendirections that the analysis concerns. Under the preconditioned flow, the evolution of the parameters are governed by
\begin{equation}
\dot{\theta} = -\frac{2}{BN}\, \Pi \sum_{\mu=1}^{B}
J_\mu^\top e^\mu,
\label{eq:param-flow}
\end{equation}
and by the chain rule each sample's error evolves as~\cite{jacot2018ntk,lee2019wide}
\begin{equation}
\dot{e}^{\nu} = J_\nu \dot{\theta}
= -\frac{2}{BN} \sum_{\mu} J_\nu\, \Pi\, J_\mu^\top e^\mu
= -(K e)^{\nu},
\qquad
K_{\nu\mu} = \tfrac{2}{BN}\, J_\nu\, \Pi\, J_\mu^\top,
\label{eq:stacked-dynamics}
\end{equation}
where the $B$ per-sample errors are stacked into one vector $e = (e^1; \dots; e^B) \in \R^{BN}$ and the per-sample Jacobians into $J := (J_1; \dots; J_B) \in \R^{BN \times P}$. We call $K = \tfrac{2}{BN}\, J \Pi J^\top$ the ``stacked kernel''. Setting $\Pi = I$ recovers the gradient-flow NTK. Written out over the batch,
\begin{equation}
\frac{\mathrm{d}}{\mathrm{d}t}
\begin{pmatrix}
e^{1} \\ e^{2} \\ \vdots \\ e^{B}
\end{pmatrix}
= -\,\frac{2}{BN}
\begin{pmatrix}
J_1 \Pi J_1^\top & J_1 \Pi J_2^\top & \cdots & J_1 \Pi J_B^\top \\[4pt]
J_2 \Pi J_1^\top & J_2 \Pi J_2^\top & \cdots & J_2 \Pi J_B^\top \\[4pt]
\vdots & \vdots & \ddots & \vdots \\[4pt]
J_B \Pi J_1^\top & J_B \Pi J_2^\top & \cdots & J_B \Pi J_B^\top
\end{pmatrix}
\begin{pmatrix}
e^{1} \\ e^{2} \\ \vdots \\ e^{B}
\end{pmatrix},
\label{eq:single-sample-view}
\end{equation}
and the $\nu$-th row is
\begin{equation}
\dot{e}^{\nu}
= -\,\frac{2}{BN}\,
\underbrace{J_\nu \Pi J_\nu^\top\, e^{\nu}}_{\text{per-sample block}}
\;-\;\frac{2}{BN}
\underbrace{\sum_{\mu \neq \nu} J_\nu \Pi J_\mu^\top\, e^{\mu}}_{\text{cross terms}} .
\label{eq:cross-terms-row}
\end{equation}
The off-diagonal blocks $J_\nu \Pi J_\mu^\top$ with $\nu \neq \mu$ are the cross terms. They couple every sample's error to every other sample's error through the shared parameters. They are absent from the block-diagonal kernel $\diag_\mu(J_\mu \Pi J_\mu^\top)$, which is what one obtains by giving each sample its own copy of the parameters, and diagnostics built on the per-sample kernel alone omit them by construction. Section~\ref{sec:error-loading} shows that dense networks have training error along the eigendirections through these cross terms.

The stacked kernel is symmetric and positive semidefinite, since $\Pi$ is diagonal with positive entries and $K = \tfrac{2}{BN} (J \Pi^{1/2}) (J \Pi^{1/2})^\top$. It therefore has an orthonormal eigenbasis $\{\phi_\alpha\}$ with eigenvalues $\lambda_\alpha \geq 0$,
\begin{equation}
K = \sum_{\alpha} \lambda_\alpha\,
\phi_\alpha \phi_\alpha^\top,
\label{eq:stacked-eig}
\end{equation}
and while the kernel is frozen the stacked error evolves diagonally in this basis,
\begin{equation}
e(t) = \sum_\alpha e^{-\lambda_\alpha t}
\left( \phi_\alpha^\top e(0) \right) \phi_\alpha.
\label{eq:eigen-decay}
\end{equation}
Each $\lambda_\alpha$ is the rate at which training contracts the error pattern $\phi_\alpha$; in the discrete iteration with learning rate $\eta$ the coefficient on $\phi_\alpha$ contracts by $1 - \eta \lambda_\alpha$ per step. The figures plot this per-step rate $\eta\lambda_\alpha$, with $\eta$ the learning rate at the sampled step. The eigenpairs $(\lambda_\alpha, \phi_\alpha)$ are the modal coordinates of the learning problem. For two-layer networks in the kernel regime, Arora et al.~\cite{arora2019fine} decompose the training error in the same way, along the eigenvectors of the Gram matrix of the network at initialization; here the kernel is that of the late-training Adam update.

Equation~\eqref{eq:eigen-decay} holds when $K$ is fixed. $K$ depends on
$\theta$ and, through $\Pi$, on the second-moment state of Adam, so it changes
over training, and the kernel at initialization does not govern the trained
error. We therefore evaluate $K$ late in training, where the loss has reached
its floor and $\Pi$ is nearly stationary over the sampled window
(Appendix~\ref{app:details}), and compare it with the error at that stage. The decomposition of the trained error over the eigendirections of $K$ in
Section~\ref{sec:error-loading} uses $K$ averaged over the final $100$ steps,
and the modal-kernel diagnostics of Section~\ref{sec:fourier-gap} use $K$ at
the final step, with $\Pi$ taken from the optimizer state at each sampled step.
% ---------------------------------------------------------------
\subsection{Error accumulates in the slow eigendirections of the kernel}
\label{sec:error-loading}
% ---------------------------------------------------------------

Because $\{\phi_\alpha\}$ is orthonormal, the trained error at a given step decomposes exactly as
\begin{equation}
e = \sum_\alpha a_\alpha \phi_\alpha,
\qquad
a_\alpha = \phi_\alpha^\top e,
\qquad
\|e\|^2 = \sum_\alpha a_\alpha^2.
\label{eq:loadings}
\end{equation}
The coefficient $a_\alpha$ is the loading of the error on eigendirection $\alpha$, and $a_\alpha^2$ is that direction's share of the squared error. We call the set of pairs $\{(\lambda_\alpha, a_\alpha^2)\}$ the error-loaded spectral measure. It represents where on the rate axis the error resides. We summarize it by the cumulative error mass below a rate $\lambda$, in RMS units,
\begin{equation}
\Gamma(\lambda) =
\Bigg( \frac{1}{BN} \!\!\sum_{\alpha:\, \lambda_\alpha < \lambda}
\!\! a_\alpha^2 \Bigg)^{\!1/2},
\label{eq:cumulative-mass}
\end{equation}
which increases from $0$ to the full RMS error $E := \|e\| / \sqrt{BN}$ as $\lambda$ increases, where $B$ is the batch size and $N$ is the number of grid points. The measure is computed by error-loaded Lanczos applied matrix-free to the late-training kernel, with the cross terms never dropped (Appendix~\ref{app:details}); the pairs $(\lambda_\alpha, a_\alpha^2)$ it returns are Ritz pairs of the kernel, which serve in place of the exact eigenpairs throughout. We locate the bulk of the measure by the rate at which the cumulative error mass reaches $95\%$ of the RMS error,
\begin{equation}
\lambda^{*} := \min \big\{ \lambda : \Gamma(\lambda) \geq 0.95\, E \big\},
\label{eq:lambda-star}
\end{equation}
so that $0.95^2 \approx 90\%$ of the squared error resides below $\lambda^{*}$. The threshold of $95\%$ is a convenience for analyzing the figures, and the conclusions do not depend on its value.

\begin{figure}[tp]
\centering
\includegraphics[width=\textwidth]{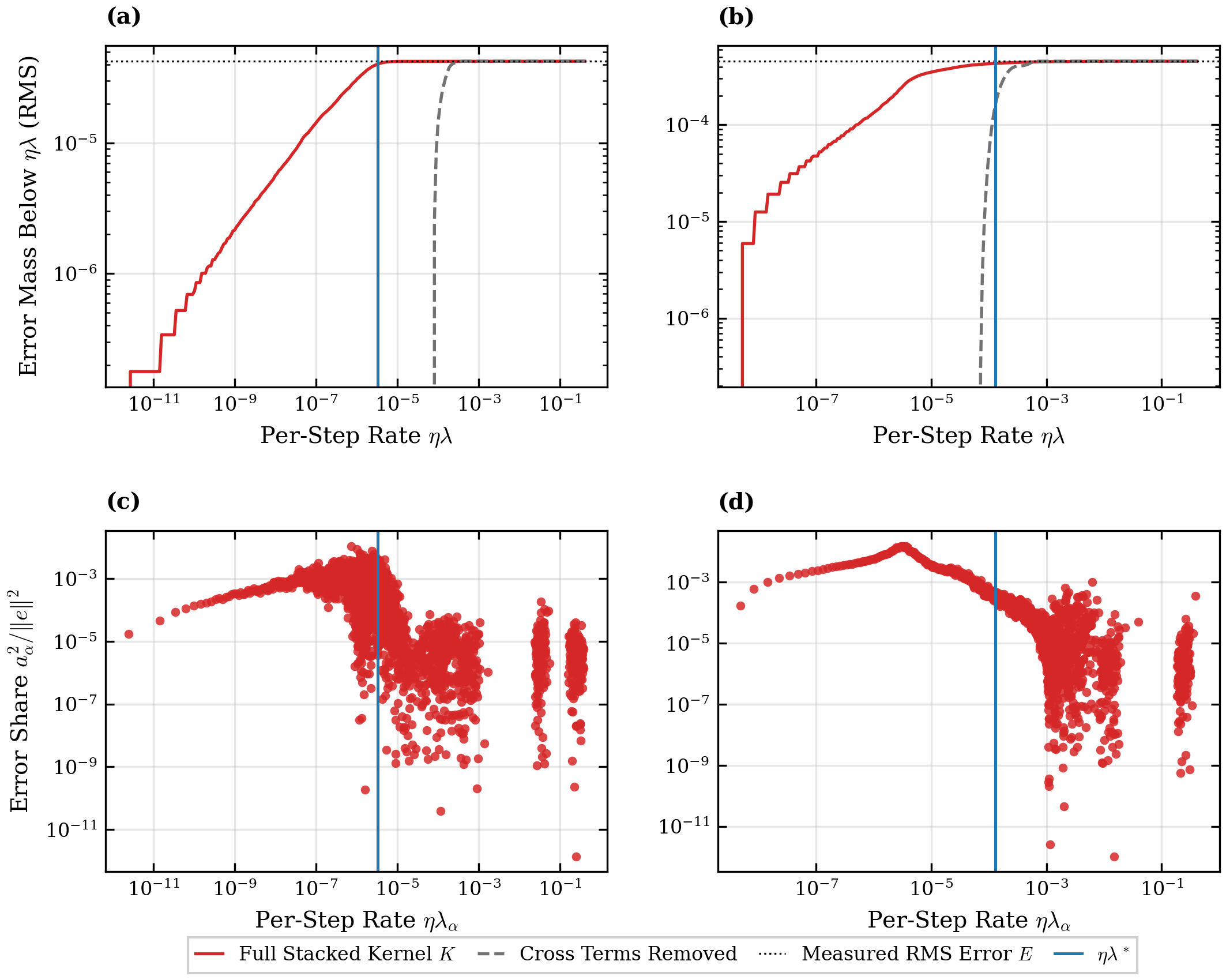}
\caption{\textbf{The error is retained in the slow eigendirections of the stacked kernel, which for this model exist only through the cross terms} (GELU dense network; smooth forcing left, white forcing right). (a,b)~Cumulative error mass $\Gamma(\lambda)$ of Eq.~(\ref{eq:cumulative-mass}) for the full stacked kernel $K$ (red) and for the block-diagonal kernel with the cross terms removed (grey dashed). The dotted line marks the measured RMS error $E$ and the vertical line marks $\eta\lambda^{*}$, with $\lambda^{*}$ of Eq.~(\ref{eq:lambda-star}) the rate at which $\Gamma$ reaches $0.95\,E$, below which about $90\%$ of the squared error resides. (c,d)~Each Ritz pair plotted at its per-step rate $\eta\lambda_\alpha$ against its error share $a_\alpha^{2}/\lVert e\rVert^{2}$.}
\label{fig:tail-flatness}
\end{figure}

The central result of this section is in Fig.~\ref{fig:tail-flatness}, which shows the measure for the GELU dense network under smooth and white forcing. For the full stacked kernel, the cumulative error mass accumulates over multiple small eigenvalues and reaches the independently measured error $E$, the dotted line in Fig.~\ref{fig:tail-flatness}(a,b). When the cross terms are removed and $K$ is replaced by its block-diagonal part, the cumulative mass collapses, leaving less than $0.1\%$ of the squared error below $\eta\lambda \approx 8 \times 10^{-5}$ for either forcing. For this model, the slow eigendirections that retain the error exist only through the coupling between samples. Panels~(c,d) resolve the same eigensystem direction by direction. Under both forcings, most of the trained error resides in the eigendirections of the stacked kernel with small eigenvalues.

The same statement holds for every model that does not learn the operator, whatever the source of the slow rates. Table~\ref{tab:hierarchy} collects the ten cases, with the training loss of each from Fig.~\ref{fig:loss-curves}. The deep linear dense network and the linear FNO under white forcing are the
two cases whose training loss reaches machine precision or is still descending
at the end of training. In neither case does the trained error reside in
eigendirections with small eigenvalues. The residual of the linear FNO
occupies two eigendirections near $\lambda^*$, and the residual of the deep
linear dense network is at machine precision and lies mostly in the null space of the kernel, with the rest at the fast end of its spectrum (Appendix~\ref{app:hierarchy-tail}). The other eight cases plateau, and in every one of them the dominant error shares lie below $\lambda^{*}$.

The table identifies three sources of slow rates. First, the forcing spectrum provides them in the smooth forcing case,  for linear and nonlinear models alike. This is the result of Section~\ref{sec:linear-op}, and it is why the deep linear dense network retains its error under smooth forcing and is at machine precision under white forcing. Second, the nonlinearity of the network provides them even when the data excite every mode. This is why the nonlinear models plateau under white forcing while the linear models learn the operator. For the GELU dense network these slow directions exist only through the cross terms: the block-diagonal kernel retains none of its error below $\eta\lambda = 10^{-5}$, against $60\%$ for the full kernel in Fig.~\ref{fig:tail-flatness}(b), so an analysis that drops the cross terms cannot locate its error. For the nonlinear FNO the cross terms also shape the slow tail: $90\%$ of the squared error lies below $\lambda^{*}$ for the full kernel and $52\%$ without the cross terms, and its slowest directions, $\eta\lambda < 10^{-7}$, exist only through them; see Table~\ref{tab:hierarchy} and Fig.~\ref{fig:tail-all}(a). For every generic model, an analysis that drops the cross terms misplaces the error on the rate axis. Third, the representation provides them as the null space of the kernel, which is the truncation of the ideal diagonal spectral operator at $k = 32$. The forcing spectrum affects the retained error in two ways. It decides which
eigenvalues are small and therefore how much error remains. It also decides
where the gap mode of Eq.~\eqref{eq:kgap} falls, because the spectrum $U_k$ of
the true solution decreases faster with $k$ under smooth forcing than under
white forcing, so the true solution drops below the error level at a lower
wavenumber. Appendix~\ref{app:hierarchy-tail} discusses each model.

\begin{table}[t]
\centering
\footnotesize
\setlength{\tabcolsep}{3pt}
\begin{tabular}{llllll}
\toprule
Model & Forcing & Loss at the end & Error-carrying directions & \begin{tabular}[t]{@{}l@{}}Error below $\lambda^{*}$\\without cross terms\end{tabular} & Fig. \\
\midrule
GELU dense & smooth & plateau & slow tail & absent (0) & \ref{fig:tail-flatness}(a,c) \\
GELU dense & white & plateau & slow tail & absent (0.12) & \ref{fig:tail-flatness}(b,d) \\
Nonlinear FNO & smooth & plateau & slow tail & present (0.49) & \ref{fig:tail-all-smooth}(a,b) \\
Nonlinear FNO & white & plateau & slow tail & present (0.53) & \ref{fig:tail-all}(a,b) \\
Linear FNO & smooth & plateau & slow tail & present (0.13) & \ref{fig:tail-all-smooth}(c,d) \\
Linear FNO & white & descending & two directions near $\lambda^{*}$ & 0.22 & \ref{fig:tail-all}(c,d) \\
Deep linear dense & smooth & plateau & slow tail & absent (0.004) & \ref{fig:tail-all-smooth}(e,f) \\
Deep linear dense & white & machine epsilon & null space and fast directions & 1.0 & \ref{fig:tail-all}(e,f) \\
Ideal diagonal spectral & smooth & truncation floor & null space & 1 (null space) & \ref{fig:tail-all-smooth}(g,h) \\
Ideal diagonal spectral & white & truncation floor & null space & 1 (null space) & \ref{fig:tail-all}(g,h) \\
\bottomrule
\end{tabular}
\caption{The hierarchy under both forcings. Training loss is the end of training from Fig.~\ref{fig:loss-curves}. The last two columns are from the error-loaded spectral measure in the figure listed: where the error resides on the rate axis, and the share of the squared error that remains below $\lambda^{*}$ when the cross terms are removed.}
\label{tab:hierarchy}
\end{table}

% ---------------------------------------------------------------
\subsection{Misalignment with the sine basis and the accuracy gap}
\label{sec:fourier-gap}
% ---------------------------------------------------------------

Spectral bias is reported in the sine basis, and what the retained error looks like there depends on how the slow eigendirections project onto the sine modes. If the slow eigendirections are individual sine modes, the retained error is localized at particular wavenumbers and each mode has its own learning rate. If they are delocalized over the sine modes, the error they retain is spread evenly over $k$. The ideal diagonal spectral operator is sine-aligned by construction, the FNOs are aligned to the extent that the forcing separates their rates, and the dense networks are indistinguishable from random directions in the sine basis under both forcings. For the dense networks the retained error is therefore flat in $k$, and measured against the decaying target spectrum this flat error reproduces the accuracy gap of Fig.~\ref{fig:spectral-gap}.

\paragraph{Alignment statistics.}
The alignment is measured on the late-training batch-averaged kernel $\bar{K} := \tfrac{2}{N}\bar{J}\, \Pi\, \bar{J}^\top \in \R^{N \times N}$, where $\bar{J}$ is the Jacobian of the batch-averaged network output on a fixed diagnostic batch (Appendix~\ref{app:details}), normalized so that $\eta\lambda_\alpha$ is the per-step rate at which an error shared by every sample of the batch contracts. $\bar{K}$ is the $N \times N$ analogue of the per-sample block $J_\mu \Pi J_\mu^\top$ with the batch average taken inside the Jacobian, and $S^\top \bar{K} S$ is its modal kernel. If the modal kernel were diagonal, the error at each sine mode would contract independently at the rate given by the corresponding diagonal entry, and spectral bias would reduce to a per-mode learning rate on the diagonal. For a unit eigenvector $\phi_\alpha$ of $\bar{K}$, let $\psi_{k\alpha} := v_k^\top \phi_\alpha$ be its component on sine mode $k$, so that $\sum_k \psi_{k\alpha}^2 = 1$. We call $\psi_{k\alpha}^2$ the squared overlaps of eigenvector $\alpha$ with the sine modes and summarize them by two statistics,
\begin{equation}
\bar{k}_\alpha := \sum_k k\, \psi_{k\alpha}^2,
\qquad
\mathrm{PR}_\alpha := \Big( \sum_k \psi_{k\alpha}^4 \Big)^{-1},
\label{eq:centroid-pr}
\end{equation}
the frequency centroid and the participation ratio of eigendirection $\alpha$. $\mathrm{PR}_\alpha = 1$ means the direction is a single sine mode. The sum $\sum_k \psi_{k\alpha}^4$ is the inverse participation ratio: it equals $1$ for a single sine mode and $1/n$ for a direction spread evenly over $n$ modes. $\mathrm{PR}_\alpha \gg 1$ means the single rate $\lambda_\alpha$ is shared across many sine modes at once, and $\mathrm{PR}_\alpha$ counts the effective number of modes over which the direction is spread.

\begin{figure}[tp]
\centering
\includegraphics[width=\textwidth]{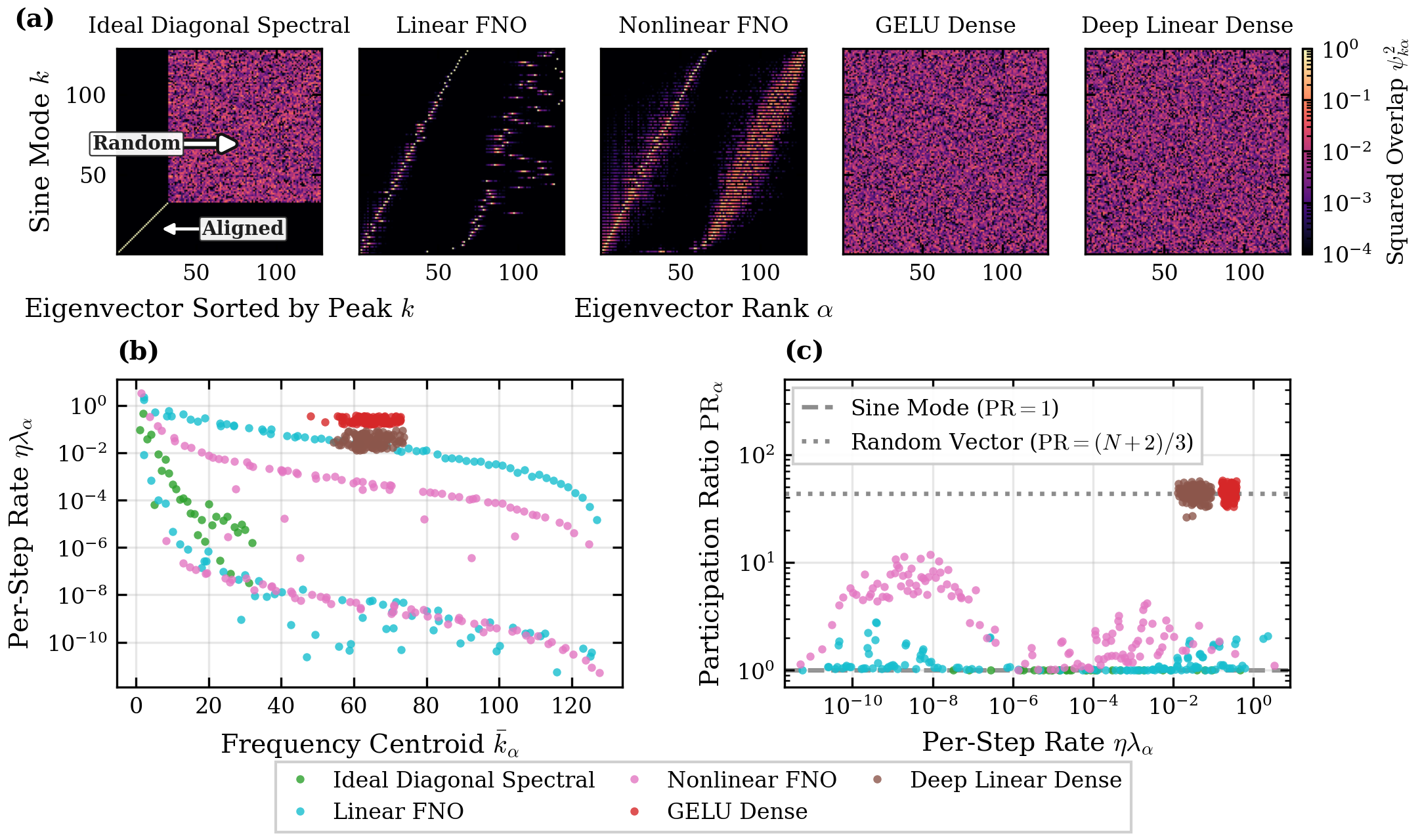}
\caption{\textbf{Alignment of the kernel eigensystem with the sine basis} (smooth forcing; late-training batch-averaged kernel $\bar{K}$ of Section~\ref{sec:fourier-gap}). (a)~Squared overlaps $\psi^2_{k\alpha}$ of Eq.~(\ref{eq:centroid-pr}) between the eigenvectors of $\bar{K}$ and the sine modes, for the five models. Eigenvectors are sorted by eigenvalue, except for the ideal diagonal spectral operator, whose retained eigenvectors are sorted by their peak sine mode $k$; its non-retained modes span the null space, whose basis is arbitrary. (b)~Per-step rate $\eta\lambda_\alpha$ of each eigenvector against its frequency centroid $\bar{k}_\alpha$. (c)~Participation ratio $\mathrm{PR}_\alpha$ of each eigenvector against its per-step rate. The dashed line marks a single sine mode, $\mathrm{PR} = 1$. The dotted line marks the expected participation ratio of a unit vector drawn uniformly at random from the sphere in $\mathbb{R}^N$, $(N+2)/3$.}
\label{fig:misalignment}
\end{figure}

We plot in Fig.~\ref{fig:misalignment}(a) the squared overlaps $\psi^2_{k\alpha}$ of
Eq.~\eqref{eq:centroid-pr}, the fraction of each eigenvector $\phi_\alpha$
of $\bar K$ that lies along sine mode $v_k$, under smooth forcing. For the ideal diagonal spectral operator every eigenvector is a single sine mode, by construction. Each of its parameters controls one output mode, its Jacobian columns are the scaled sine modes $\hat{f}^\mu_k\, v_k$, and its modal kernel is diagonal at every forcing. The diagonal entries of $S^\top \bar{K} S$ scale with $\langle \hat{f}_k \rangle^2$, the batch average of the forcing coefficients, so under smooth forcing they follow $\sigma_k^2 = k^{-4}$ and the rate falls with $k$ in Fig.~\ref{fig:misalignment}(b). The non-retained modes span the null space of the kernel, so representational truncation is absorbed by the same law as the $\lambda_\alpha \to 0$ limit. The linear FNO is sine-diagonal, with a single stripe of squared overlaps equal to one ordered by rank. The nonlinear FNO is banded around a stripe, with each eigenvector localized on a few sine modes within one parity sector by the selection rule discussed in Appendix~\ref{app:parity}. The GELU dense and deep linear dense networks show no sine structure at all. Each of their eigenvectors spreads over many sine modes, so each rate is shared across many modes at once.

Figure~\ref{fig:misalignment}(b) plots each eigenvalue against the frequency centroid of its eigenvector. For the ideal diagonal spectral operator and the FNOs, the rate falls as the centroid increases, by about five orders of magnitude over the $32$ retained modes of the ideal operator and by ten orders of magnitude from $0$ to $N$ for the FNOs. Error in eigendirections with more high-wavenumber content therefore decays
more slowly, which is spectral bias represented in the eigenbasis of the kernel.

The decrease of $\lambda_\alpha$ with $\bar k_\alpha$ follows from the forcing
spectrum $\sigma_k^2 = k^{-4}$, which appears in the modal kernel through the
Jacobian columns. Adam divides gradient coordinate $i$ by $s_i^{1/2} + \epsilon_{\mathrm{A}}$.
Where $s_i^{1/2} \gg \epsilon_{\mathrm{A}}$ this division cancels the scale
of the gradient on parameter $i$ and would remove the forcing spectrum from
the kernel. By the end of training, the gradients on most parameters have decayed below $\epsilon_{\mathrm{A}}$: $s_i^{1/2} < \epsilon_{\mathrm{A}}$ for between $93.5\%$ and $99.9\%$ of the parameters of the GELU dense network and the nonlinear FNO under both forcings, and for all of the FNO spectral weights on modes $k > 64$ under smooth forcing. On those parameters,
$\Pi_{ii} \approx \epsilon_{\mathrm{A}}^{-1}$ and the kernel reduces to
$\tfrac{2}{BN}\,\epsilon_{\mathrm{A}}^{-1} J J^\top$, in which the forcing spectrum
appears through the Jacobian columns (Appendix~\ref{app:hierarchy-tail}). The
ordering of the eigenvalues with $\bar k_\alpha$ is therefore not removed by
the preconditioner. The kernel makes high-wavenumber directions slow to the extent that the training data fail to excite them.

For the linear FNO, the eigenvectors are the sine modes and the rates follow the forcing spectrum, so its learning order is the data covariance through the kernel, as for the linear model of Section~\ref{sec:linear-op}. For the nonlinear FNO, the nonlinearity mixes neighboring modes, and the learning order depends on the data through the kernel but is no longer the covariance alone. The dense networks instead collapse to clusters near $\bar{k}_\alpha \approx N/2$ with little spread in rate. The centroid of a delocalized eigenvector is close to the middle of the band regardless of its rate, so the centroid carries no information for these models. Figure~\ref{fig:misalignment}(c) gives the participation ratios. Eigenvectors on the random-vector line are statistically indistinguishable from random directions in the sine basis. The dense networks lie on this line, the linear FNO at $\mathrm{PR} = 1$ at every rate, and the nonlinear FNO between $1$ and about $10$, with its largest values at its slowest rates.

Note that a Frobenius measure of diagonality, i.e., the fraction of the squared entries of the modal kernel $S^\top \bar{K} S$ that lie on the diagonal, does not detect this misalignment. The eigenvectors of a matrix are determined by the size of its off-diagonal entries relative to the gaps between its diagonal entries, so when many diagonal entries are nearly equal, off-diagonal entries that are negligible in Frobenius norm are still large enough to mix the corresponding modes, and the eigenvectors become arbitrary combinations of the near-degenerate sine modes. This is the situation for the GELU dense network, whose kernel is close to a multiple of the identity matrix and is therefore nearly diagonal in the sine basis by the Frobenius measure, even though its eigenvectors are indistinguishable from random directions in that basis, as Fig.~\ref{fig:misalignment}(c) shows. A per-mode learning rate exists only if the eigenvectors are single sine modes ($\mathrm{PR} = 1$) and the corresponding rates differ from mode to mode. This holds for the ideal diagonal spectral operator and for the smooth-forcing linear FNO but not for the other models. This is why the analysis of Section~\ref{sec:error-loading} uses the eigenbasis of the kernel rather than the sine basis.

\begin{figure}[tp]
\centering
\includegraphics[width=\textwidth]{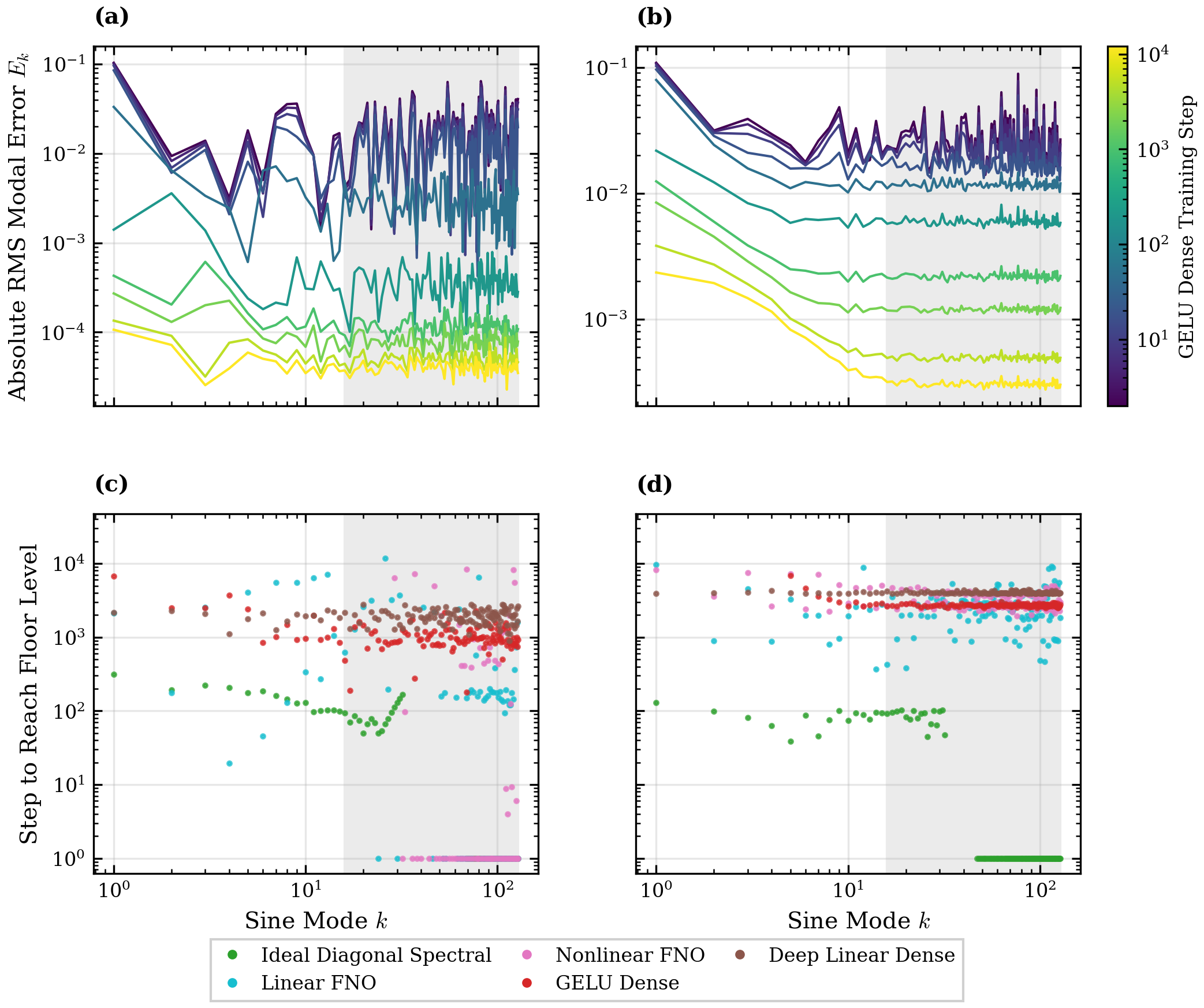}
\caption{\textbf{The modal error of a delocalized neural operator is flat in $k$ and every mode reaches its floor at the same time} (smooth forcing left, white forcing right). (a,b)~Modal error $E_k$ of the GELU dense network at logarithmically spaced training steps. The shaded band marks the gap region ($k \ge k_{\mathrm{gap}}$). (c,d)~Training step at which the error of each mode first reaches the floor level, $3\times$ the final gap-median error, for every model. Modes whose error lies below this level from the first step are plotted at step $1$.}
\label{fig:phenomenon-operator}
\end{figure}
\paragraph{The flat floor is the projection of the slow eigendirections.}
We now return to the sine basis, in which spectral bias is reported. Figure~\ref{fig:phenomenon-operator}(a,b) shows the modal error $E_k$ of the GELU dense network at logarithmically spaced training steps under smooth and white forcing. For both forcings, beyond a low-$k$ regime, the trained spectrum is flat in $k$ and descends uniformly as training proceeds, i.e., training lowers the error at all wavenumbers at the same rate. This is explained by the kernel eigensystem. A slow, delocalized eigendirection contributes the same small residual to every mode onto which it projects. Let $\mathcal{A} := \{\alpha : \lambda_\alpha < \lambda^{*}\}$ be the set of slow eigendirections of Eq.~(\ref{eq:lambda-star}) and $e_{\mathrm{slow}} := \sum_{\alpha \in \mathcal{A}} a_\alpha \phi_\alpha$ the part of the error they retain. For a stacked eigendirection $\phi_\alpha \in \R^{BN}$ with per-sample blocks $\phi_\alpha^\mu$, write $\psi_{k\alpha}^2 := \sum_\mu (v_k^\top \phi_\alpha^\mu)^2$ for its squared overlap with sine mode $k$, which again sums to one, over all $k$. The trained modal error is then
\begin{equation}
E_k^2 \;\approx\; \frac{1}{B} \sum_{\alpha \in \mathcal{A}}
\psi_{k\alpha}^2\, a_\alpha^2
\;\approx\; \frac{\| e_{\mathrm{slow}} \|^2}{BN},
\label{eq:flat-floor}
\end{equation}
where the factor $1/B$ is the average over the $B$ samples in the RMS of Eq.~(\ref{eq:abs-modal-error}), and the second approximation uses $\psi_{k\alpha}^2 \sim 1/N$ for directions with $\mathrm{PR}_\alpha = \mathcal{O}(N)$, as in Fig.~\ref{fig:misalignment}(c), since a direction spread over $\mathcal{O}(N)$ modes places weight of order $1/N$ on each of them. The result is independent of $k$ and, in particular, independent of the true solution at mode $k$. The measured errors $E$ of the diagnostic batch marked in Fig.~\ref{fig:tail-flatness} and Figs.~\ref{fig:tail-all} and~\ref{fig:tail-all-smooth} coincide with the flat levels of the same models on the validation set in Fig.~\ref{fig:final-spectra}, so the kernel diagnostics describe the same error that Fig.~\ref{fig:final-spectra} reports. Against the decaying target spectrum $U_k \approx \sigma_k/\kappa_k$, this flat floor produces the gap mode of Eq.~(\ref{eq:kgap}) and the growing relative error of Fig.~\ref{fig:final-spectra}, as shown in Section~\ref{sec:phenomenon}, with $\varepsilon_k = E_k/U_k \propto \kappa_k/\sigma_k$, which grows as $k^4$ under smooth forcing and as $k^2$ under white forcing for $k \ll N$, as the slope triangles in Fig.~\ref{fig:final-spectra}(c,d) show. The growth of $\varepsilon_k$ beyond $k_{\mathrm{gap}}$ is therefore not evidence that the network fails at high wavenumbers. It is an error that is the same at every wavenumber, measured against a target whose spectrum decays through it. The higher the floor, the earlier the gap: the model with the largest $E$ under smooth forcing, the deep linear dense network, departs from the target first in Fig.~\ref{fig:spectral-gap}(e).

The linear FNO under smooth forcing is the opposite case, with the error
arranged mode by mode rather than flat. Its kernel is sine-diagonal there
, as Fig.~\ref{fig:misalignment}(c) shows, so the error on each sine mode decays
independently at its own rate, as in Eq.~\eqref{eq:linear-op-sine}, and the
high modes, which smooth forcing rarely excites, are learned last. Its error is
jagged in Fig.~\ref{fig:final-spectra}(a) because each eigendirection retains
its error at its own wavenumber rather than spreading it by
Eq.~\eqref{eq:flat-floor}. The deep linear dense network retains high-mode error for the same reason, but its eigenvectors are random in the sine basis under both forcings, as Figs.~\ref{fig:misalignment}(c) and~\ref{fig:misalignment-white}(c) show, and its floor is flat. The nonlinear FNO lies between them. Its slowest eigendirections have $\mathrm{PR}$ of order ten and there are many of them across the band, so their sum is almost flat (Fig.~\ref{fig:final-spectra}, pink).

Over a window in which the kernel is close to fixed, such as the closing window in which $K$ is sampled, the time evolution of the error follows from the same decomposition; it does not describe the whole run of Fig.~\ref{fig:phenomenon-operator}, over which the kernel changes. While the
kernel is fixed, $a_\alpha(t) = a_\alpha(0)\,e^{-\lambda_\alpha t}$ by
Eq.~\eqref{eq:eigen-decay}, and substituting into
Eq.~\eqref{eq:abs-modal-error} gives
\begin{equation}
E_k^2(t) = \frac{1}{B}\sum_{\alpha,\beta} a_\alpha(0)\,a_\beta(0)\,
e^{-(\lambda_\alpha+\lambda_\beta)t}\sum_\mu
(v_k^\top\phi_\alpha^\mu)(v_k^\top\phi_\beta^\mu).
\label{eq:survival-full}
\end{equation}
The terms with $\alpha = \beta$ are
\begin{equation}
\sum_\alpha \chi_{k\alpha}\, e^{-2\lambda_\alpha t},
\qquad \chi_{k\alpha} = \psi_{k\alpha}^2\, a_\alpha^2(0),
\label{eq:survival}
\end{equation}
and the terms with $\alpha \neq \beta$ do not vanish, since the eigendirections
are orthogonal in $\R^{BN}$ but their projections onto a single sine mode are
not orthogonal over the samples. At late times both sums are dominated by the
eigendirections with the smallest $\lambda_\alpha$ that have nonzero weight on
mode $k$. Note that every mode couples weakly to some fast directions through the weights $\chi_{k\alpha}$, so an arithmetic average of the eigenvalues acting on a mode is dominated by the fast end of the spectrum and does not predict the final error. The eigenvalues must be exponentiated before they are averaged, and the final error of a mode is governed by the slowest eigendirections to which it couples.

\paragraph{Synchronous floor times.}
Figure~\ref{fig:phenomenon-operator}(c,d) restates the uniform descent of Fig.~\ref{fig:phenomenon-operator}(a,b) mode by mode. It shows the training step at which the error of each mode first reaches its floor level, under smooth and white forcing. Within the gap, the dense networks, whose slow eigendirections are delocalized, reach the floor at essentially the same step across all modes under both forcings, whereas the FNOs are spread over an order of magnitude in training step. Mode-indifferent learning is a property of the delocalized architectures, not of the problem or of the forcing.

\paragraph{Error level and error shape are determined separately.}
The eigenvalues and eigenvectors of the stacked kernel appear in Eq.~(\ref{eq:survival}) through different factors. The eigenvalues govern how much error is retained during training, the error mass loaded below $\lambda^{*}$ (Fig.~\ref{fig:tail-flatness}; Figs.~\ref{fig:tail-all} and~\ref{fig:tail-all-smooth}). The eigenvectors determine how that mass is arranged over $k$, through the squared overlaps $\psi^2_{k\alpha}$ shown in Eq.~(\ref{eq:flat-floor}). Under white forcing, the deep linear dense network and the linear FNO, one random in the sine basis and one localized in Fig.~\ref{fig:misalignment-white}(c), both learn the operator in Fig.~\ref{fig:loss-curves}(b) and have no accuracy gap in Fig.~\ref{fig:spectral-gap}(f). Under smooth forcing, the same deep linear dense network, with the same random projection in Fig.~\ref{fig:misalignment}(c), retains error in the slow tail, and a gap in accuracy is observed in Figs.~\ref{fig:tail-all-smooth}(e,f) and~\ref{fig:spectral-gap}(e). The GELU dense network and the nonlinear FNO, random and localized in the same way, both retain error in the slow tail under white forcing, and a gap in accuracy is observed. Under smooth forcing, all four generic models retain error at comparable levels, with $E = 3.6 \times 10^{-5}$ for the linear FNO, $3.3 \times 10^{-5}$ for the nonlinear FNO, $4.2 \times 10^{-5}$ for the GELU dense network and $2.6 \times 10^{-4}$ for the deep linear dense network, and their spectra range from rising and jagged for the linear FNO to flat for the dense networks according to their projection in Fig.~\ref{fig:final-spectra}(a). Whether spectral bias appears is decided by the eigenvalues. The projection onto the sine basis decides its shape, which is flat for delocalized directions by Eq.~(\ref{eq:flat-floor}) and mode by mode for sine-aligned directions.

% ===============================================================
\section{Discussion}
\label{sec:discussion}
% ===============================================================

In the cases studied here, the error evolves under a linear or linearized update, the modal rates are the eigenvalues of the operator that governs that update, and any statement about wavenumbers is a statement about how the eigenvectors of that operator project onto the sine basis. In shallow linear operator learning the operator is the forcing covariance $\Sigma_f$ (Section~\ref{sec:linear-op}). In nonlinear operator learning it is the stacked kernel of the Adam update, whose eigenstructure depends on the architecture through the Jacobians, on the data through the forcing, and on the optimizer through the preconditioner $\Pi$ (Section~\ref{sec:nonlinear}). In fixed-basis scalar regression (Appendix~\ref{sec:scalar}) it is the Gram matrix $G = \Phi^\top \Phi$ of the basis on the sample grid, and spectral bias vanishes exactly when the sampling makes $G$ proportional to the identity (Appendix~\ref{sec:orthogonal}).

This viewpoint resolves the apparent conflict with classical numerical analysis. Weighted Jacobi iteration for the same Poisson problem damps oscillatory error fastest because its update operator $I - \omega\, \diag(A)^{-1} A$, with $\omega = 2/3$, is built from the Poisson operator, whose largest eigenvalues belong to the high modes, so the oscillatory half of the spectrum is reduced by at least a factor of three per iteration while smooth error persists~\cite{briggs2000multigrid}. Adam training of the ideal diagonal spectral operator and the FNOs learns the low wavenumbers first under smooth forcing for a different reason. For these models the eigenvalue of the stacked kernel decreases as the wavenumber content of its eigenvector increases in Fig.~\ref{fig:misalignment}(b), so error in directions with more high-wavenumber content decays more slowly. For the dense networks the eigenvectors carry no wavenumber information and no such ordering exists. Neither ordering is universal. Each is the spectrum of a different operator in the same basis, and the ordering changes when the operator changes, whether through the architecture, the data, or the algorithm.

The analysis restricts what spectral bias can mean for generic architectures. Whether a per-mode learning rate exists is determined by the projection. The dense networks never have one, since their eigenvectors are statistically indistinguishable from random directions in the sine basis under both forcings. The FNOs have one to the extent that the forcing separates their rates, fully for the linear FNO under smooth forcing and only weakly under white forcing. For a sine-aligned kernel the classical description is correct, and the high wavenumbers are learned slowly because the data rarely excite them. For a delocalized kernel the observation usually called spectral bias, a relative error growing with wavenumber, is the combination of two facts, neither of which is a mode-by-mode law. Error retained in the slow eigendirections of the kernel produces an error floor that is flat across the sine modes, and the spectrum of the true solution decays through that floor. Where the data spectrum is broad the gap opens late and the apparent bias is mild. Where it decays steeply the gap opens early and the apparent bias is severe. Eigenvector alignment with the sine basis therefore cannot by itself predict spectral bias. It predicts the form the error takes once the eigenvalues have fixed its size, and spectral bias is a wavenumber-selective learning rule only when that projection is sine-aligned.

% \section*{Author contributions}

\section*{Acknowledgments}
C.-Y.L. acknowledges the NSF via grant OAC-2608400. AC and CA acknowledge the support from the National Science Foundation (grant no. 2425667), DARPA APaQus program, the Sloan Foundation, and Schmidt Sciences, LLC. A.C. and C.-Y.L. designed the research. C.A. wrote the code and executed the research. Early versions of the code were provided by A.C. A.C. and C.A. wrote the paper, and all the authors analyzed the results and edited the manuscript.

\section*{Data and code availability}
All data in this study are generated by the code: training forcings are sampled from the ensembles of Section~\ref{sec:poisson-setup} and the targets are computed exactly. The code for training the models, computing the kernel diagnostics and reproducing every figure is available on GitHub at \url{https://github.com/cainsliewastaken/Spectral_Learning}.

\bibliographystyle{unsrt}
\bibliography{references}

\appendix
\setcounter{figure}{0}
\renewcommand{\thefigure}{A\arabic{figure}}

% ===============================================================
\section{Spectral bias in scalar regression}
\label{sec:scalar}
% ===============================================================

In scalar regression, whether spectral bias is well posed depends on the operator that governs the update and on how many coordinates it acts on.

% ---------------------------------------------------------------
\subsection{One trainable coefficient}
\label{sec:shallow}
% ---------------------------------------------------------------

Let the model be $y_\theta(x) = w x$ with $w \in \R$. Given training data $\{(x_i, y_i)\}_{i=1}^{n_{\mathrm{s}}}$, the empirical squared loss is
\begin{equation}
\mathcal{L}(w) = \frac{1}{2} \sum_{i=1}^{n_{\mathrm{s}}} \left( w x_i - y_i \right)^2,
\qquad
S_{xx} := \sum_{i=1}^{n_{\mathrm{s}}} x_i^2,
\qquad
S_{xy} := \sum_{i=1}^{n_{\mathrm{s}}} x_i y_i,
\label{eq:moments}
\end{equation}
with the unique minimizer $w^{*} = S_{xy}/S_{xx}$ provided $S_{xx} > 0$. Gradient descent with learning rate $\eta > 0$ gives
\begin{equation}
w^{n+1} - w^{*} = \left( 1 - \eta S_{xx} \right)
\left( w^{n} - w^{*} \right),
\label{eq:single-mode}
\end{equation}
which converges linearly if $0 < \eta < 2 / S_{xx}$. Equation~(\ref{eq:single-mode}) is the complete learning dynamics of the model. The model class is one-dimensional, so there is one rate and no decomposition into modes learned at different rates. Trained on an oscillatory target, the model converges to the least-squares projection of the target onto the span of $x$. This failure comes from the representation, not from the training dynamics. In the operator-learning setting the same failure appears as the null space of the kernel ($\lambda_\alpha = 0$, Section~\ref{sec:error-loading}).

% ---------------------------------------------------------------
\subsection{Depth}
\label{sec:deep-linear}
% ---------------------------------------------------------------

\begin{figure}[t]
\centering
\includegraphics[width=\textwidth]{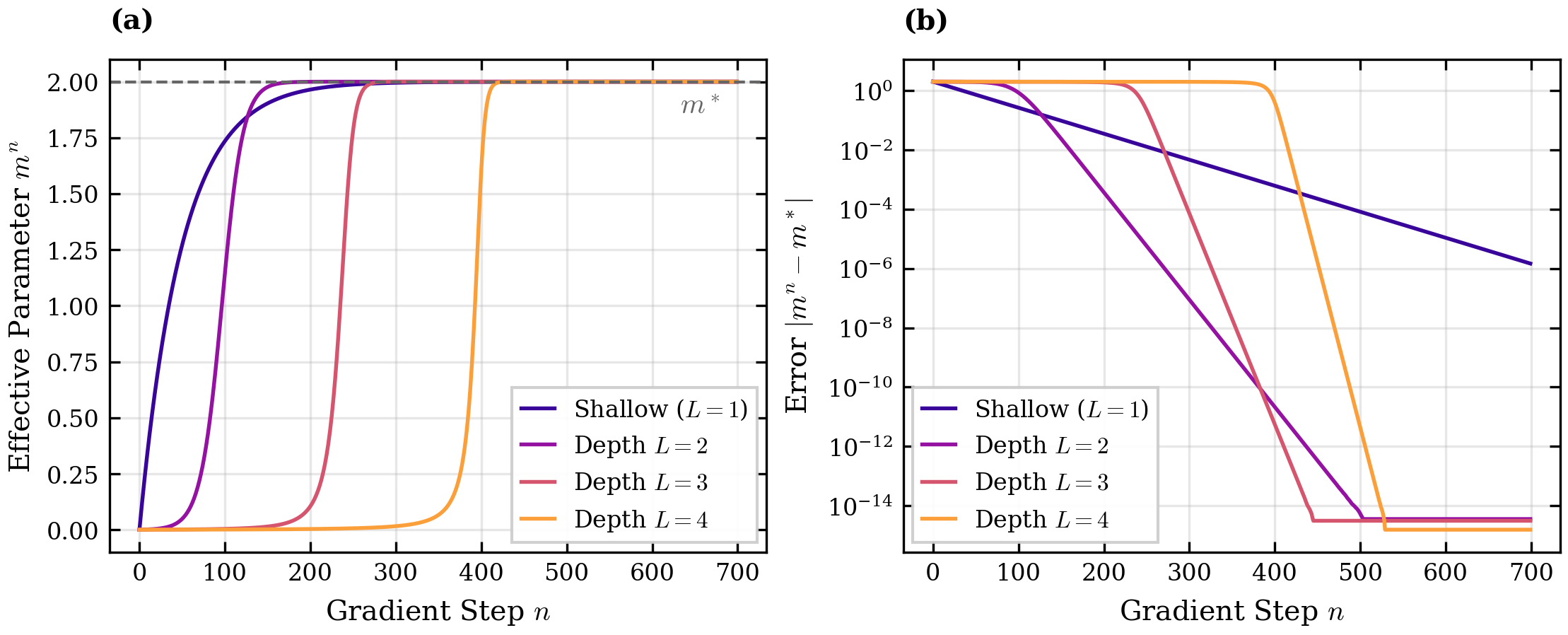}
\caption{\textbf{Depth changes the optimization, not the function class}. All depths represent the same map $x \mapsto m x$ and converge to the same target $m^*$, but the trajectory of the effective parameter $m^n$ (a) and the error $|m^n - m^*|$ (b) depend on the factorization: the shallow model decays as a single geometric mode, whereas deeper factorizations exhibit an extended plateau followed by a rapid transition, governed by Eq.~(\ref{eq:effective-flow}).}
\label{fig:depth}
\end{figure}

A depth-$L$ linear network $y_\theta(x) = W_L W_{L-1} \cdots W_1 x$ with scalar input and output represents $x \mapsto m x$ for an effective scalar $m$, so depth does not enlarge the function class; it changes the optimization. For the factorized scalar model
\begin{equation}
y_\theta(x) = \Big( \prod_{\ell=1}^{L} w_\ell \Big) x = m x,
\qquad
m := \prod_{\ell=1}^{L} w_\ell ,
\label{eq:factorized}
\end{equation}
the loss is the quadratic of Eq.~(\ref{eq:moments}) in $m$. Gradient flow on the factors, $\dot{w}_\ell = -\partial \mathcal{L} / \partial w_\ell$ with $\partial m / \partial w_\ell = m / w_\ell$, induces
\begin{equation}
\dot{m}
= - \left( S_{xx} m - S_{xy} \right)
   m^2 \sum_{\ell=1}^{L} \frac{1}{w_\ell^2}.
\label{eq:effective-flow}
\end{equation}
The convergence speed depends on the factor magnitudes through $m^2 \sum_{\ell} w_\ell^{-2}$~\cite{saxe2014exact,arora2018optimization}. For small initialization, deeper factorizations produce long plateaus followed by rapid transitions to the target, as Fig.~\ref{fig:depth} shows. The function class is still $\{x \mapsto m x\}$, so there is still no notion of spectral bias.

% ---------------------------------------------------------------
\subsection{Prescribed sine modes}
\label{sec:sine-basis}
% ---------------------------------------------------------------

Suppose the target is $u^{*}(x) = \sum_{k=1}^{n_{\mathrm{m}}} c_k \sin(k \pi x)$ and the model uses the same basis with trainable amplitudes, $u_\theta(x) = \sum_{k=1}^{n_{\mathrm{m}}} a_k \sin(k \pi x)$. With the design matrix $\Phi \in \R^{n_{\mathrm{s}} \times n_{\mathrm{m}}}$, $\Phi_{ik} = \sin(k \pi x_i)$, on the training points $x_1, \dots, x_{n_{\mathrm{s}}}$, the empirical loss is $\mathcal{L}(\bm{a}) = \tfrac{1}{2} (\bm{a} - \bm{c})^\top \Phi^\top \Phi\, (\bm{a} - \bm{c})$, and gradient descent with step size $\eta$ gives
\begin{equation}
\bm{a}^{n+1} - \bm{c}
= \left( I - \eta\, G \right) (\bm{a}^{n} - \bm{c}),
\qquad
G := \Phi^\top \Phi.
\label{eq:coeff-dynamics}
\end{equation}
The learning dynamics are governed entirely by the Gram matrix $G$. This is the analog, in coefficient space, of the scalar contraction factor of Eq.~(\ref{eq:single-mode}), with $n_{\mathrm{m}}$ trainable spectral coordinates in place of one.

% ---------------------------------------------------------------
\subsection{Orthogonal sampling}
\label{sec:orthogonal}
% ---------------------------------------------------------------

If the training grid makes the sine basis orthogonal on the sample points, then $\Phi^\top \Phi = \gamma I$ for some $\gamma > 0$ and each coefficient evolves independently,
\begin{equation}
a_k^{n+1} - c_k = (1 - \eta \gamma)\, (a_k^{n} - c_k),
\qquad k = 1, \dots, n_{\mathrm{m}}.
\label{eq:equal-rates}
\end{equation}
Every represented frequency is learned at exactly the same rate. There is neither low-frequency-first nor high-frequency-first learning, as Fig.~\ref{fig:modal-rates}(a) shows.

% ---------------------------------------------------------------
\subsection{Non-orthogonal sampling}
\label{sec:non-orthogonal}
% ---------------------------------------------------------------

\begin{figure}[t]
\centering
\includegraphics[width=\textwidth]{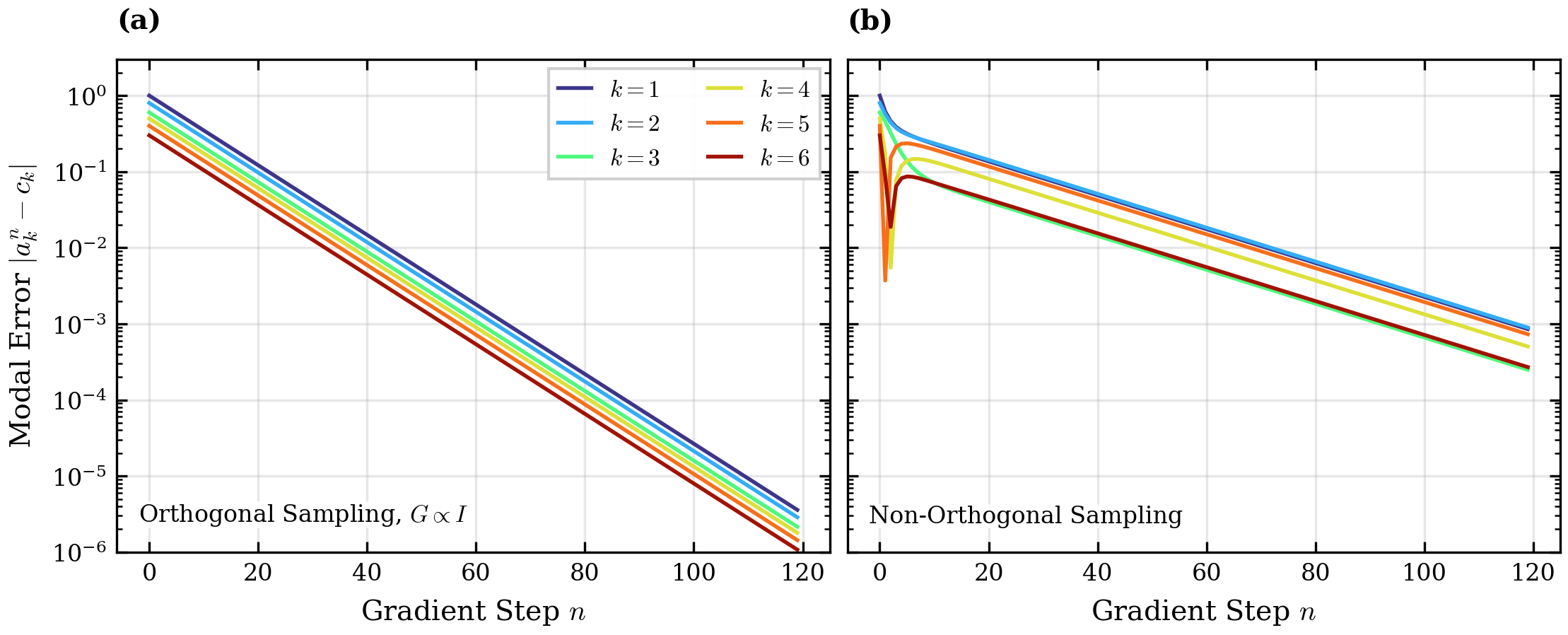}
\caption{\textbf{Fixed sine-basis model: spectral order is governed by the data geometry through the Gram matrix}. (a)~Under orthogonal sampling, the sine modes diagonalize $G$ and every modal error $|a_k^n - c_k|$ contracts at an identical rate, Eq.~(\ref{eq:equal-rates}). (b)~Under non-orthogonal sampling the sine modes are coupled by Eq.~(\ref{eq:coeff-dynamics}); individual sine coefficients exhibit transient non-monotone behavior, and the exponential rates act on the eigenvectors of $G$, Eq.~(\ref{eq:eigenbasis-rates}).}
\label{fig:modal-rates}
\end{figure}

If the sample points do not make the sine basis orthogonal, $G$ is not diagonal and Eq.~(\ref{eq:coeff-dynamics}) mixes the coefficients. With $G = O\, \Lambda_G\, O^\top$, $\Lambda_G = \diag(\mu_1, \dots, \mu_{n_{\mathrm{m}}})$ and $O$ orthogonal, the components $\xi^n := O^\top(\bm{a}^{n} - \bm{c})$ evolve as
\begin{equation}
\xi_j^{\,n+1} = (1 - \eta \mu_j)\, \xi_j^{\,n}.
\label{eq:eigenbasis-rates}
\end{equation}
The rates are the eigenvalues of $G$, and they act on its eigenvectors, not on the sine modes. Individual sine coefficients are coupled and can be transiently non-monotone, as Fig.~\ref{fig:modal-rates}(b) shows. Section~\ref{sec:nonlinear} has the same structure, with the stacked kernel in place of $G$.

% ===============================================================
% ===============================================================
\section{Architectures and training details}
\label{app:details}
% ===============================================================

All experiments use the discrete Poisson problem of Section~\ref{sec:poisson-setup} with $N = 128$ interior grid points, in double precision. Training pairs are generated fresh at every optimization step by sampling forcing coefficients as in Section~\ref{sec:poisson-setup} and solving $u = A^{-1} f$ exactly.

\paragraph{Ideal diagonal spectral operator.}
The model is
\begin{equation}
u_\theta = S\, \Lambda_\theta\, S^\top f,
\qquad
\Lambda_\theta = \diag(g_1, \dots, g_{K_{\mathrm{ret}}}, 0, \dots, 0),
\label{eq:ideal-model}
\end{equation}
with one trainable gain per retained sine mode and no other parameters, $K_{\mathrm{ret}} = 32$ retained modes, and parameters initialized at zero. Its Jacobian columns are the scaled sine modes $\hat{f}^\mu_k\, v_k$, so its modal kernel is exactly diagonal in the sine basis. It is the $\mathrm{PR} = 1$ reference of the hierarchy.

\paragraph{Fourier neural operators.}
The FNOs lift the grid signal to $d_{\mathrm{c}} = 32$ channels with a pointwise linear layer, apply $D = 3$ FNO layers, and project back to one channel with a pointwise linear layer. Each FNO layer sums a pointwise linear channel-mixing branch and a spectral branch that transforms to sine coefficients, applies an independent trainable $d_{\mathrm{c}} \times d_{\mathrm{c}}$ weight to each retained mode, and transforms back. The runs retain all $128$ modes. The nonlinear FNO applies a GELU after each layer and the linear FNO applies none. Spectral weights are initialized from a zero-mean Gaussian with standard deviation $0.05$. The lift, local, and projection layers carry biases, whose spatially constant output underlies the parity selection rule of Appendix~\ref{app:parity}.

\paragraph{Dense networks.}
The GELU dense network is a multilayer perceptron acting on the $128$-dimensional grid vector, of depth $4$ and width $512$, with GELU activations between layers. The deep linear dense network has the same depth and width, no nonlinearities, and no bias terms, with weights initialized from a zero-mean Gaussian with standard deviation $0.05$. Its product of weight matrices realizes a single linear operator, so it is the operator-learning analogue of the deep linear scalar model of Appendix~\ref{sec:deep-linear}.

\paragraph{Training.}
All models are trained with Adam ($\beta_1 = 0.9$, $\beta_2 = 0.999$, $\epsilon_{\mathrm{A}} = 10^{-8}$) on the mean-squared error, with batch size $512$, freshly sampled forcings at every step, gradient-norm clipping at $1$, and $12000$ optimization steps. The learning rate is $10^{-2}$ for the ideal diagonal spectral operator and $5 \times 10^{-5}$ for all other models, annealed by a cosine schedule to $1\%$ of its initial value over the run. Error spectra are computed on a fixed validation set of $4096$ forcings per ensemble.

\paragraph{Kernel diagnostics.}
Except where noted, kernel statistics in the main text are computed over the final $100$ optimization steps and averaged, so they describe the operator that drives the update at the stage of training whose error is measured. The stacked kernel $K = \tfrac{2}{BN} J \Pi J^\top$ is applied matrix-free as $z \mapsto \tfrac{2}{BN} J \Pi (J^\top z)$ for $z \in \R^{BN}$, with $\Pi$ of Eq.~(\ref{eq:adam-precond}) assembled from the second-moment state of Adam at the sampled step and with exact automatic-differentiation Jacobian-vector products. The cross terms enter through the sum over samples and are never dropped. The error-loaded spectral measure of Eq.~(\ref{eq:cumulative-mass}) is computed by error-loaded Lanczos~\cite{golub2010matrices,lin2016approximating}. Initializing the Krylov recursion at $e$ yields the Ritz values $\lambda_\alpha$ and the weights $a_\alpha^2$ directly, with full reorthogonalization. The recursion is run for $m = 2000$ steps on the $BN$-dimensional stacked kernel, and stops earlier only when the Krylov space is exhausted, as it is for the linear models whose error occupies a few directions. The $m$ Ritz pairs form the $m$-point Gauss quadrature of the error-loaded spectral measure~\cite{golub2010matrices}: the weights sum to $\|e\|^2$ exactly, $\Gamma(\lambda)$ of Eq.~(\ref{eq:cumulative-mass}) is evaluated on them, and each Ritz pair stands for the group of exact eigendirections it resolves, so the per-direction panels of Figs.~\ref{fig:tail-flatness}, \ref{fig:tail-all} and~\ref{fig:tail-all-smooth} show Ritz pairs. The block-diagonal measure needs no Lanczos: each per-sample block $J_\mu \Pi J_\mu^\top$ is diagonalized exactly, for $512$ samples of each batch. Reported measures are averaged over four independent forcing batches of size $B = 2048$. The modal-kernel diagnostics of Section~\ref{sec:fourier-gap} use the batch-averaged kernel $\bar{K} = \tfrac{2}{N}\bar{J}\, \Pi\, \bar{J}^\top$, with $\bar{J}$ the exact Jacobian of the batch-averaged network output on a fixed diagnostic batch of $32$ forcings and $\Pi$ assembled at the final optimization step. All modal projections use the sine basis of Eq.~(\ref{eq:sine-basis}).

\paragraph{Validity of the diagonal Adam model.}
The diagonal form of $\Pi$ is not an approximation. The second-moment state of Adam is elementwise by construction, so Eq.~(\ref{eq:adam-precond}) is the preconditioner the optimizer applies. The two modeling steps in Section~\ref{sec:kernel-bias} are the neglect of momentum and the use of a fixed $\Pi$ at each sampled step. Momentum averages the gradient over roughly $(1-\beta_1)^{-1}$ steps. For an eigendirection with $\eta \lambda_\alpha \ll 1 - \beta_1$ the error is quasi-static over that memory, the averaged gradient equals the instantaneous one, and the momentum-free model returns the same rate. Momentum modifies only the fast end of the spectrum, $\eta \lambda_\alpha \gtrsim 1 - \beta_1$. The error-carrying rates lie below $\eta\lambda^{*} \le 1.3 \times 10^{-4}$ for every model that plateaus, three orders of magnitude below $1 - \beta_1 = 0.1$ (Figs.~\ref{fig:tail-flatness}, \ref{fig:tail-all} and~\ref{fig:tail-all-smooth}). The fast ends of the linear models' spectra, where $\eta \lambda_\alpha$ reaches order one, carry no error beyond the machine-precision residual of the deep linear dense network. This argument concerns the rate at which each direction contracts within the sampled window. It does not describe how momentum distributed the error over the eigendirections earlier in training. The second-moment state moves on the timescale of $(1-\beta_2)^{-1} = 1000$ steps, ten times the final $100$-step window, over which the trained error remains at its floor and the gradients are noise-dominated. $\Pi$ is therefore nearly stationary across the window and the bias corrections are close to one. Gradient-norm clipping at $1$ is applied at every step. It does not act in the sampled window: at the final weights the pre-clip gradient norm is below $2 \times 10^{-5}$ for every model, so the clipped and unclipped updates coincide there. In this regime, Eq.~(\ref{eq:stacked-dynamics}) is the linearization of the actual update.

% ===============================================================
\section{The hierarchy under white and smooth forcing}
\label{app:hierarchy-tail}
% ===============================================================

\begin{figure}[!p]
\centering
\includegraphics[width=\textwidth,height=0.83\textheight,keepaspectratio]{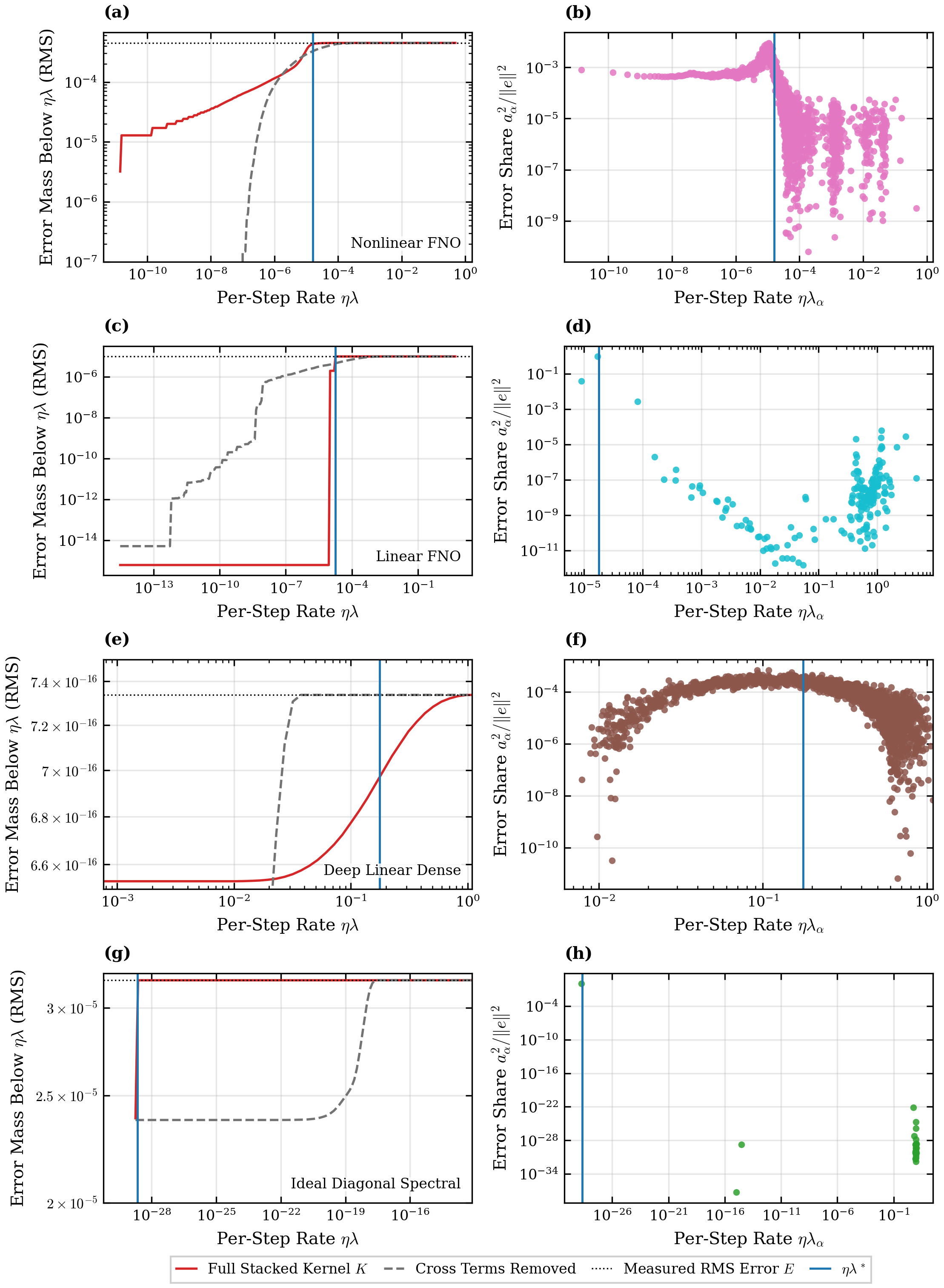}
\caption{\textbf{The error-loaded spectral measure of Fig.~\ref{fig:tail-flatness}, repeated for the other four models} (white forcing; rows are nonlinear FNO (a,b), linear FNO (c,d), deep linear dense (e,f), and ideal diagonal spectral (g,h)). Left column, cumulative error mass $\Gamma(\lambda)$ of Eq.~(\ref{eq:cumulative-mass}) for the full stacked kernel (red) and for the block-diagonal kernel with the cross terms removed (grey dashed), with the measured RMS error $E$ (dotted) and $\eta\lambda^{*}$ with $\lambda^{*}$ of Eq.~(\ref{eq:lambda-star}) (vertical line). Right column, each Ritz pair at its per-step rate $\eta\lambda_\alpha$ against its error share $a_\alpha^{2}/\lVert e \rVert^{2}$. The rate axes differ across rows (Appendix~\ref{app:hierarchy-tail}).}
\label{fig:tail-all}
\end{figure}

\begin{figure}[t]
\centering
\includegraphics[width=\textwidth]{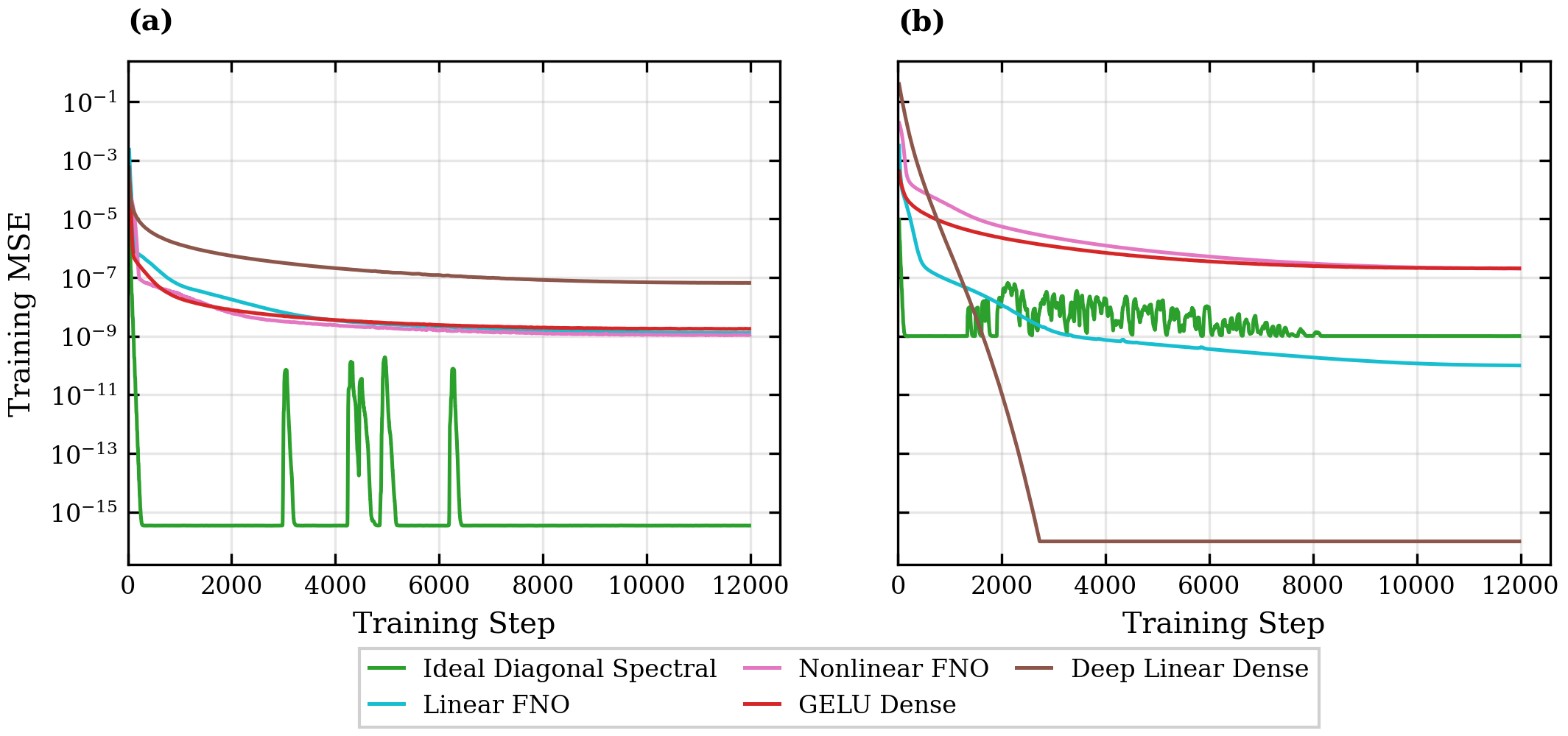}
\caption{\textbf{Training loss of the hierarchy} under smooth (a) and white (b) forcing.}
\label{fig:loss-curves}
\end{figure}

Figure~\ref{fig:tail-flatness} reported the error-loaded spectral measure of Eq.~(\ref{eq:cumulative-mass}) for the GELU dense network. Figure~\ref{fig:tail-all} repeats the diagnostic under white forcing for the other four models, with rows ordered as nonlinear FNO (a,b), linear FNO (c,d), deep linear dense (e,f), and ideal diagonal spectral (g,h). Figure~\ref{fig:loss-curves} shows the training loss of every model under both forcings and identifies the level to which each model learns the operator. The comparison separates the hierarchy into two groups. The nonlinear models (nonlinear FNO, Fig.~\ref{fig:tail-all}(a,b); GELU dense, Fig.~\ref{fig:tail-flatness}) plateau at a training loss far above machine epsilon under white forcing in Fig.~\ref{fig:loss-curves}(b) and reproduce the structure of Section~\ref{sec:error-loading}. The linear models learn the operator under white forcing, the deep linear dense network to machine epsilon by roughly step $4000$ and the linear FNO to a loss still descending at the end of training in Fig.~\ref{fig:loss-curves}(b), and their measures in Fig.~\ref{fig:tail-all}(c--f) carry none of that structure. Under white forcing the slow tail of Section~\ref{sec:error-loading} is therefore a property of the nonlinear operators rather than of Adam training on this problem in general. Under smooth forcing the forcing spectrum supplies slow rates to every model, as discussed below.

\paragraph{Nonlinear models.}
For the nonlinear FNO in Fig.~\ref{fig:tail-all}(a,b), the full stacked kernel behaves as in Section~\ref{sec:error-loading}. Its cumulative error mass rises over rates from $\eta\lambda \approx 10^{-11}$ to $\eta\lambda^{*}$ and reaches the measured error $E$, and the dominant error shares are carried by the eigendirections below $\lambda^{*}$ in Fig.~\ref{fig:tail-all}(b), the tail that produces the flat floor through Eq.~(\ref{eq:flat-floor}). The cross terms shape the slow tail here as well. Removing them eliminates the slowest directions, $\eta\lambda < 10^{-7}$, which carry $2\%$ of the squared error, and moves part of the mass between $\eta\lambda \approx 10^{-5}$ and $\eta\lambda^{*}$ to rates up to $\eta\lambda \approx 10^{-4}$, leaving $52\%$ of the squared error below $\lambda^{*}$ against $90\%$ for the full kernel in Fig.~\ref{fig:tail-all}(a).

\paragraph{Linear models.}
The linear models depart from this picture in both columns. For the linear FNO the full-kernel measure is nearly a single step in Fig.~\ref{fig:tail-all}(c). A single eigendirection at $\lambda_\alpha \approx \lambda^{*}$ carries the dominant share of the squared error, a second direction just below $\lambda^{*}$ carries nearly all of the remainder, and the directions above $\lambda^{*}$ carry shares smaller by several orders of magnitude in Fig.~\ref{fig:tail-all}(d). Removing the cross terms redistributes this mass over rates from below $\eta\lambda = 10^{-7}$ up to $\eta\lambda \approx 10^{-3}$ without destroying it, and the block-diagonal measure reaches the same observed error. For the deep linear dense network the trained error is at machine epsilon ($\epsilon_{\mathrm{mach}} \approx 2.2 \times 10^{-16}$), so the measure characterizes a machine-epsilon residual rather than unlearned operator structure. That residual is not of the form $\Delta f$ for a linear map $\Delta$, so it lies mostly outside the range of the kernel: $79\%$ of its squared error sits in the numerical null space of the kernel ($\eta\lambda_\alpha \approx 10^{-18}$, below the plotted range), and the rest is spread over many directions at $\eta\lambda_\alpha \approx 10^{-2}$ to $1$ in Fig.~\ref{fig:tail-all}(e,f). For the ideal diagonal spectral operator in Fig.~\ref{fig:tail-all}(g,h), the kernel is exactly diagonal in the sine basis and the trained residual is the representational truncation of the non-retained modes. The error mass sits in the null space of the kernel, at zero rate to numerical precision, and the remaining directions carry shares smaller by many orders of magnitude in Fig.~\ref{fig:tail-all}(h).

Under white forcing, in neither linear model does removing the cross terms erase the error-carrying mass, and in neither does the error occupy a slow, wide range of rates. The three properties that produce the flat floor in Section~\ref{sec:fourier-gap}, slowness, delocalization, and existence only through the cross terms, coincide in the GELU dense network under white forcing. The nonlinear FNO shares the slowness, but its eigendirections spread over a few sine modes rather than over all of them, and the cross terms deepen its slow tail rather than create all of it. The models that learn the operator lack these properties, each in a different way. This narrows the scope of Section~\ref{sec:error-loading} to the nonlinear operators without changing the mechanism. Both linear models realize representable linear maps of a linear problem, and white forcing excites every input direction (Section~\ref{sec:linear-op}), so little residual survives training for the cross terms to retain.

\begin{figure}[!p]
\centering
\includegraphics[width=\textwidth,height=0.83\textheight,keepaspectratio]{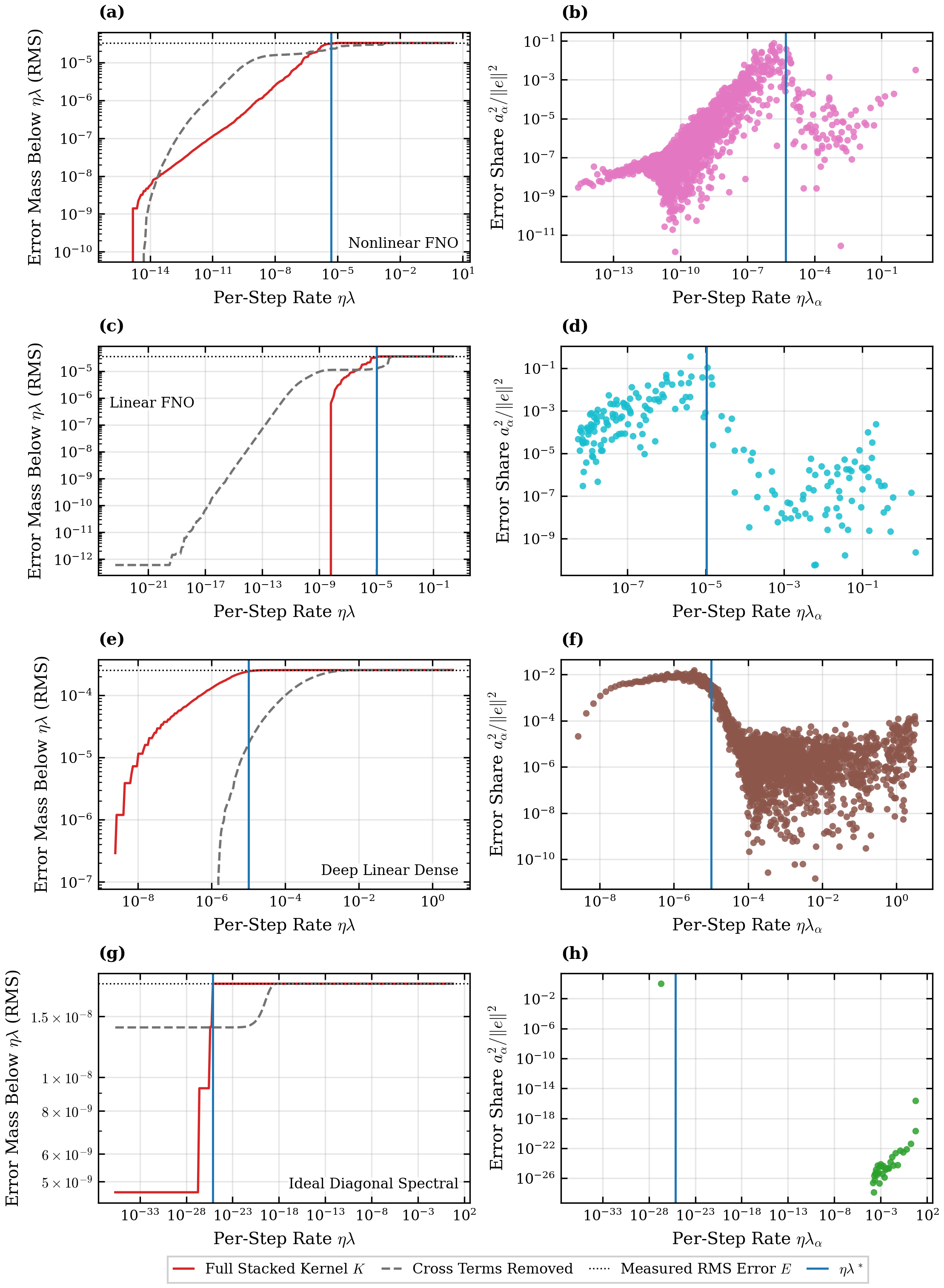}
\caption{\textbf{The error-loaded spectral measure of Fig.~\ref{fig:tail-all}, repeated under smooth forcing}. Rows are nonlinear FNO (a,b), linear FNO (c,d), deep linear dense (e,f), and ideal diagonal spectral (g,h); the GELU dense network is shown in Fig.~\ref{fig:tail-flatness}(a,c). Lines and markers as in Fig.~\ref{fig:tail-all}.}
\label{fig:tail-all-smooth}
\end{figure}

\paragraph{Smooth forcing.}
Under smooth forcing every generic model plateaus at a training loss between $10^{-9}$ and $10^{-7}$ in Fig.~\ref{fig:loss-curves}(a), and Fig.~\ref{fig:tail-all-smooth} shows that in every case the dominant error shares lie below $\lambda^{*}$. The deep linear dense network, whose residual is at machine epsilon under white forcing, now retains its error over slow rates spanning about four orders of magnitude in Fig.~\ref{fig:tail-all-smooth}(e,f), and removing the cross terms empties this range as it does for the GELU dense network. This is what the linear analysis of Section~\ref{sec:linear-op} predicts, since smooth forcing rarely excites the high modes and the error on them decays slowly. For the FNOs the block-diagonal measure also carries mass at slow rates in Fig.~\ref{fig:tail-all-smooth}(a,c), so under smooth forcing the slow tail of these models does not depend on the cross terms alone.

\begin{figure}[t]
\centering
\includegraphics[width=\textwidth]{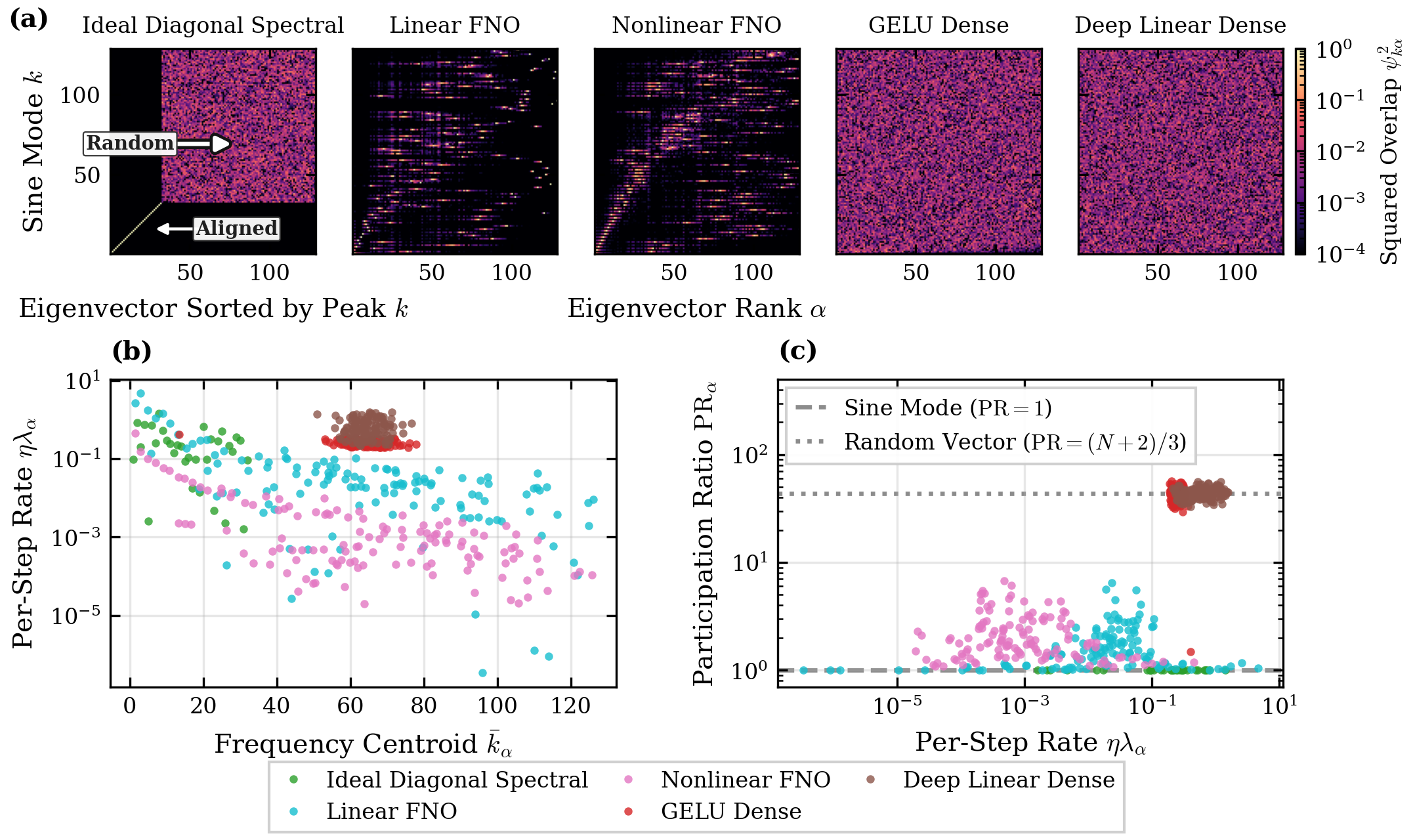}
\caption{\textbf{Alignment with the sine basis under white forcing} (late-training batch-averaged kernel $\bar{K}$; panels as in Fig.~\ref{fig:misalignment}). (a)~Squared overlaps $\psi^2_{k\alpha}$ between the eigenvectors of $\bar{K}$ and the sine modes, sorted as in Fig.~\ref{fig:misalignment}(a). (b)~Per-step rate $\eta\lambda_\alpha$ against the frequency centroid $\bar{k}_\alpha$. (c)~Participation ratio $\mathrm{PR}_\alpha$ against per-step rate, with the reference lines of Fig.~\ref{fig:misalignment}(c).}
\label{fig:misalignment-white}
\end{figure}

\paragraph{Alignment under white forcing.}
Figure~\ref{fig:misalignment-white} repeats the alignment diagnostics of Fig.~\ref{fig:misalignment} under white forcing. The dense networks are unchanged. Their overlaps carry no structure in Fig.~\ref{fig:misalignment-white}(a), their centroids cluster at $N/2$ with no spread in rate in Fig.~\ref{fig:misalignment-white}(b), and their participation ratios lie on the random-vector line in Fig.~\ref{fig:misalignment-white}(c). The FNOs differ. Their eigenvectors are localized on small sets of modes within one parity sector rather than on single modes, with participation ratios of order a few at every rate. The anti-correlation between rate and centroid in (b) is weak, with the rates spread over a few orders of magnitude rather than ten. By the argument of Section~\ref{sec:fourier-gap}, the compressed rates under white forcing leave the FNO spectra near-degenerate, and the channel mixing scrambles the eigenvectors within each near-degenerate group. For the ideal diagonal spectral operator the eigenvectors are still single sine modes, and with its retained eigenvectors sorted by peak $k$ they again form a clean diagonal in (a). Under white forcing, however, the diagonal entries of its modal kernel are determined by the zero-mean batch average of equal-variance coefficients, so its rates are only weakly ordered in $k$.

\paragraph{Eigenvalue magnitudes and the optimizer state.}
The rate axes of Fig.~\ref{fig:tail-all} span different ranges. The deep linear dense spectrum lies at $\eta\lambda_\alpha \sim 10^{-2}$ to $1$ and the linear FNO spectrum extends to $\eta\lambda_\alpha \approx 5$, while the error-carrying rates of the models that plateau lie at $\eta\lambda_\alpha \lesssim 10^{-4}$. The ordering follows from the preconditioner of Eq.~(\ref{eq:adam-precond}). Once the residual of a model is small, its gradients are small, the second-moment estimates $s_i$ decay on the $(1-\beta_2)^{-1}$-step timescale, and the entries of $\Pi = \diag\big((s_i^{1/2} + \epsilon_{\mathrm{A}})^{-1}\big)$ grow toward $\epsilon_{\mathrm{A}}^{-1}$. Since $K = \tfrac{2}{BN} J \Pi J^\top$, the late-training eigenvalues grow accordingly, which is why the models with the lowest loss have the largest kernel eigenvalues. The magnitude of $\lambda_\alpha$ at the end of training therefore measures the optimizer state as much as the Jacobian geometry. Each model should be read on its own rate axis. What can be compared across models is the shape of the measure relative to $\lambda^{*}$ and the participation ratios, not the rates themselves.

% ===============================================================
\section{A parity selection rule for the FNO kernels}
\label{app:parity}
% ===============================================================

The overlap matrices of the FNOs in Fig.~\ref{fig:misalignment}(a) carry an alternating banded structure in $k$. This is a symmetry of the problem and of the architecture, not an artifact of training. Let $R \in \R^{N \times N}$ be the reflection about the midpoint of the domain, $(R f)_j = f_{N+1-j}$, the grid form of $x \mapsto 1 - x$. The sine modes have alternating parity under $R$,
\begin{equation}
R\, v_k = (-1)^{k+1} v_k ,
\label{eq:parity}
\end{equation}
so odd-$k$ modes are reflection-even and even-$k$ modes are reflection-odd, and the discrete Poisson operator commutes with $R$. The FNOs are reflection-equivariant at every parameter value. Their spectral weights act mode by mode, and their lift, local channel mixing, projection, and pointwise nonlinearity all commute with the reflection. The local channel mixing acts identically at every grid point, so it commutes with the sine transform of each channel and couples no modes. Without biases or nonlinearity the FNO is therefore diagonal in the sine basis, and the only mode coupling in the linear FNO comes from the biases. The bias terms of these layers add a spatially constant function, which is reflection-even and therefore has sine content on odd modes only, so the dominant mode-coupling channel of the linear FNO acts within one parity sector. Equivariance gives $F_\theta(Rf) = R\, F_\theta(f)$, so the modal kernel commutes with $R$ in expectation over forcings and for a reflection-symmetric batch; an individual batch is not symmetric, and the per-sample fluctuations break the symmetry. Its dominant eigenvectors concentrate in a single parity sector, and $\psi_{k\alpha}^2 \approx 0$ for $k$ of the wrong parity. To leading order each eigenvector acts on every other mode, which is why the structure appears as bands rather than exact zeros. The dense networks have no reflection symmetry, so their diagnostics show no bands. As in the scalar case of Appendix~\ref{sec:non-orthogonal}, the modal behavior follows from which directions the kernel makes available, not from wavenumber.

The same bias terms account for the residual of the linear FNO under white forcing. Its output at $f = 0$ lies on odd modes only and carries $99.99\%$ of its squared error, almost all of it at $k = 125$ and $k = 127$. A constant has sine content at every odd mode, so the concentration at $k = 125$ and $k = 127$ is determined by the spectral weights downstream of the biases. The offset persists because its direction is the slowest of the linear FNO's kernel, as Fig.~\ref{fig:tail-all}(d) shows. These are the only modes at which its error exceeds $U_k$, and they appear as the isolated spike at the end of the band in Figs.~\ref{fig:spectral-gap}(f) and~\ref{fig:final-spectra}(b,d). At every other mode its error is of order $10^{-7}$. The mean square of the offset, about $10^{-10}$, equals the final training loss of the linear FNO in Fig.~\ref{fig:loss-curves}(b). The residual is therefore almost entirely an input-independent offset rather than an error in the learned linear map.

\end{document}